\documentclass{article}

\usepackage{amsmath}
\usepackage{amssymb}
\usepackage{graphicx}
\usepackage[preprint]{neurips_2026}
\usepackage[utf8]{inputenc} 
\usepackage[T1]{fontenc}    
\usepackage{hyperref}      
\usepackage{url}           
\usepackage{booktabs}       
\usepackage{amsfonts}       
\usepackage{nicefrac}       
\usepackage{microtype}      
\usepackage{xcolor}         
 
\title{A Mechanistic Interpretability Study of an Astronomy Foundation Model}

\author{%
  Rachel C. Zhang$^{1,2}$ \quad Alberto Bietti$^{2,3}$ \quad Wes Gurnee \\
  \textbf{Alessandro Favero$^{1,2}$ \quad Miles Cranmer$^{1,2}$} \\
  $^1$University of Cambridge \quad 
  $^2$Polymathic AI \quad
  $^3$Flatiron Institute \\
  \small Correspondence to: \texttt{rcz20@cam.ac.uk}
}

\begin{document}

\maketitle

\begin{abstract}
Foundation models are increasingly used across the sciences, but whether their internal computations correspond to known scientific operations or proceed through different mechanisms remains unclear. Understanding these mechanisms is essential for assessing reliability, identifying failure modes, and learning new science from the models. We present, to our knowledge, the first causal mechanistic interpretability study of an astronomy foundation model, AION-1, focusing on its multimodal advantage in redshift prediction, where combining imaging with photometry outperforms photometry alone. We find that this advantage is largest for spatially extended and dust-attenuated galaxies, and that most of it is localized to the first of the encoder's twelve blocks. Within that block, how well a property is linearly decodable does not predict whether it can be localized. Spatial extent is causally used, and one attention head dominates linear probing for it, yet ablating that head has negligible effect on redshift prediction: the information is distributed across heads and is decodable again one layer after we erase it. Dust attenuation shows the opposite pattern: it is only weakly decodable anywhere in the encoder, yet a single head is necessary for most of the multimodal advantage, particularly for the most dust-attenuated galaxies, and keeping that head as the only active head in its block preserves nearly all of the advantage. None of the astrophysically motivated operations we tested explains what this head computes. Whether this reflects a limitation of current interpretability methods or a computation that does not correspond to any known astrophysical operation remains open, and resolving it is a prerequisite for using interpretability as a tool for scientific discovery.

\end{abstract}

\section{Introduction} 

Foundation models trained on heterogeneous corpora are now being applied across the natural sciences, from molecular biology~\citep{rives2021esm, jumper2021alphafold} to multi-physics surrogates~\citep{mccabe2023multiplephysics} to astronomy~\citep{parker2024astroclip}, with the goal of producing models that generalize across many downstream scientific tasks. Astronomy is a particularly compelling setting for this paradigm. Its data are inherently multimodal: multi-wavelength photometry, spectra, and imaging each capture distinct information about the same object, and combining them can provide a more comprehensive understanding of astrophysical phenomena. Additionally, many properties of astrophysical interest, such as redshift and stellar mass, are physically measurable, so what a model has learned can be checked directly against them.              

The largest such effort in astronomy to date is AION-1~\citep{aion2025}, a $300\text{M}$--$3.1\text{B}$ parameter family of omnimodal models trained on more than 200~million observations from five large-scale surveys (Legacy Survey, HSC, SDSS, DESI, Gaia). It extends 4M-style multimodal masked modeling~\citep{mizrahi2023fourm} to 39 astronomical modalities in a single transformer and reports strong performance on regression, classification, generation, and retrieval. Yet despite the centrality of such models to data-driven astrophysics, almost nothing is known about how they integrate modalities or why multimodal inputs improve scientific predictions. This matters because astronomers want to use these models to discover astrophysics, not merely to predict it.

Mechanistic interpretability has begun to answer analogous questions in vision and language models by identifying features and circuits that implement specific behaviors~\citep{olah2020zoom, anthropic2025tracing, anthropic2025biology}. Applied to scientific foundation models, it can also produce actionable domain hypotheses: \citet{goodfire2026alzheimer} reverse-engineered an epigenetic disease classifier and surfaced a fragmentomic biomarker hypothesis not central to the prevailing Alzheimer's literature. These tools have known limitations, however: probes reveal what is decodable rather than what is causally used~\citep{belinkov2022probing}, and ablating a component may not remove a property the model encodes redundantly~\citep{mcgrath2023hydra}. Whether these failure modes arise in a scientific model, where ground truth can be verified numerically, has not been tested.

We apply these methods to AION-1 through a concrete, well-studied problem: photometric redshift estimation. A galaxy's redshift $z$, the fractional shift of its spectrum toward longer wavelengths due to cosmic expansion, is a proxy for its distance. It can be measured precisely from a spectrum, but spectroscopy is slow and observationally expensive to scale, so photometric methods instead infer redshift from a galaxy's brightness in a few broadband filters~\citep{salvato2019photoz}. AION-1 attains $R^2 = 0.869$ from photometric flux values only (\emph{Ph}) and $R^2 = 0.931$ when galaxy images are also provided (\emph{Ph+Im}). This multimodal advantage is not well understood and raises a natural mechanistic interpretability question: why does adding images to photometry improve redshift prediction?

We find that the galaxies whose predictions improve under \emph{Ph+Im} are systematically more spatially extended and dust-attenuated. We investigate which components of AION-1 produce the multimodal advantage, what they encode, and how their computations compare to hypotheses for how images improve redshift prediction. In particular, our work makes three main contributions:  

\begin{itemize}        
    \item \textbf{First causal mechanistic study of an astronomy foundation model.} We apply linear and nonlinear probing, attention analysis, ablation, activation and path patching, and least-squares concept erasure to address a question that black-box evaluation alone cannot answer.

    \item \textbf{Linear decodability does not imply localization.} We localize most of the multimodal advantage to AION-1's first encoder block. Within it, one head best linearly encodes spatial extent and has an attention pattern and output geometry consistent with computing it, yet the head has a negligible causal effect on redshift prediction. This is not because spatial extent is causally unused: the model relies on it, but the information is spread across heads and remains decodable after that head is removed. A strong probe therefore does not identify the component that matters, and a null ablation does not show that a property is unused. 
        
   \item \textbf{Localization of most of the multimodal advantage to a single head.} A different attention head in the same encoder block is causally necessary for most of the multimodal advantage and preserves nearly all of it when the other heads in that block are mean-ablated. On the most dust-attenuated galaxies, imaging reduces attenuation bias, the statistical compression of predictions toward the mean~\citep{moran2026regions}, and this head is required for that reduction: ablating the head leaves the prediction more attenuation-biased than the photometry baseline. To our knowledge, this is the first localization of such a reduction to a single model component. What the head computes remains unidentified: probes of its output recover only weak dust information, and none of the astrophysically motivated operations we tested predicts its behavior.  
\end{itemize}   

\section{Related work}
\paragraph{Multimodal foundation models for astronomy.}  Cross-modal pretraining was introduced to astronomy by AstroCLIP~\citep{parker2024astroclip}, which contrastively aligns galaxy images and DESI spectra and matches a supervised ResNet-18 on photometric redshifts. AstroM$^3$~\citep{rizhko2025astrom3} extends this to photometric time series, and recent multimodal masked autoencoders learn joint image--spectrum representations~\citep{mmae2025}. AION-1~\citep{aion2025} extends this lineage to omnimodal sequence modeling across 39 modalities.

\paragraph{Photometric redshift estimation and the role of images.} Photometric redshift estimation has a long history of template-fitting and supervised ML methods~\citep{salvato2019photoz}. Image-based deep learning approaches~\citep{pasquet2018cnnphotoz, henghes2022deepphotoz, merz2024deepdisc, khederlarian2026roman} consistently outperform feature-based regressors. The improvement is attributed to morphological information such as galaxy size, shape, and surface brightness~\citep{morphoz2020}. \citet{moran2026regions} observe that image-based methods suffer less attenuation bias than photometry-based ones, which they attribute to deep networks combining the redshift information in individual pixels more effectively than integrated photometry does. These accounts are correlational and do not examine what the model internally computes.
    
\paragraph{Mechanistic interpretability and the limits of probing and ablation.} Mechanistic interpretability, developed largely on language models,  has two widely used classes of methods, observational probes and causal interventions. Probing is observational: a property can be cleanly decodable from a representation without being used downstream~\citep{belinkov2022probing}. Causal interventions such as activation patching~\citep{geiger2022causal} and ablation were developed to close this gap, and recent work shows explicit dissociation between decodability and causal use in controlled settings~\citep{decodecausal2026}. On the other hand, ablation can fail to cleanly remove redundantly encoded properties. Individual units and heads can be removed from a trained network with little loss~\citep{michel2019heads, morcos2018directions}, and ablated components are frequently compensated for downstream~\citep{mcgrath2023hydra, wang2022ioi}. We show that both limitations arise in a scientific foundation model.

\paragraph{Interpretability for astronomy models.} Existing interpretability work on astronomy models characterizes learned features but does not test their causal role post hoc. Sparse-by-design architectures~\citep{wu2025sfnet} ensure causal use of features by construction, and post-hoc sparse autoencoders on pretrained encoders~\citep{wu2025euclid} recover monosemantic morphology features without verifying causal use. Closer to our setting, \citet{dey2022capsule} train a capsule network to estimate photometric redshifts and identify a feature hypothesized to distinguish intrinsically red ellipticals from dust-reddened edge-on spirals. We follow a post-hoc approach and add causal interventions.

\section{Setup} 
\subsection{AION-1 and the redshift prediction task}
\label{sec:setup-aion}

We use the AION-1 base model (${\sim}300$M parameters) from Polymathic AI~\citep{aion2025} with the encoder frozen. The encoder has 12 transformer blocks with hidden dimension 768, each with 12 attention heads of dimension 64. Following AION-1's downstream regression setup, we train a redshift \emph{readout}, consisting of a lightweight cross-attention pooling layer and two-layer MLP, under MSE loss on the frozen encoder output to predict the spectroscopic redshift $z$  (architecture, training hyperparameters, and robustness checks in Appendix~\ref{app:redshift_pred}).  

We compare two input configurations and train one readout for each. \textbf{Photometry (\emph{Ph})} uses nine scalars: a galaxy's total flux in four optical Legacy Survey bands ($g, r, i, z$) and four infrared WISE bands ($W1$--$W4$), plus the Galactic foreground extinction $E(B{-}V)$, a measure of Milky Way dust along the line of sight.\footnote{The AION-1 paper uses only the four Legacy Survey fluxes for \emph{Ph}. We follow the documented nine-scalar specification, which strengthens the \emph{Ph} baseline relative to AION-1's reported numbers.} Each scalar is mapped to a single discrete token by its own learned codec, giving nine encoder tokens. \textbf{Photometry + Image (\emph{Ph+Im})} adds a four-band $g, r, i, z$ DESI Legacy Imaging Survey cutout ($160{\times}160$ pixels)~\citep{dey2019legacysurvey}, the spatially resolved counterpart of the optical fluxes. AION-1's image codec encodes the cutout as a $24{\times}24$ grid of 576 vector-quantized patch tokens, for 585 encoder tokens in total.           

\subsection{Data}      
\label{sec:setup}
We use the DESI Early Data Release galaxy sample~\citep{desi2024edr}, distributed in the Multimodal Universe format~\citep{mmu2024}. This dataset cross-matches Legacy Survey imaging and photometry~\citep{dey2019legacysurvey} with DESI spectroscopic redshifts and with physical properties such as stellar mass and dust attenuation inferred by PROVABGS~\citep{hahn2023provabgs}, which fits stellar-population models to each galaxy's spectral energy distribution. We restrict the dataset to the Bright Galaxy Survey (BGS) bright sample with $z_{\rm spec} > 0$, finite $g,r,i,z$ Legacy Survey fluxes, valid magnitudes in $g, r, z, W1$, and a positive PROVABGS stellar mass. This yields $67{,}755$ galaxies, which we split $80/20$ into train and test sets ($54{,}204$ / $13{,}551$).

\begin{figure}[h]
    \centering
    \includegraphics[width=\textwidth]{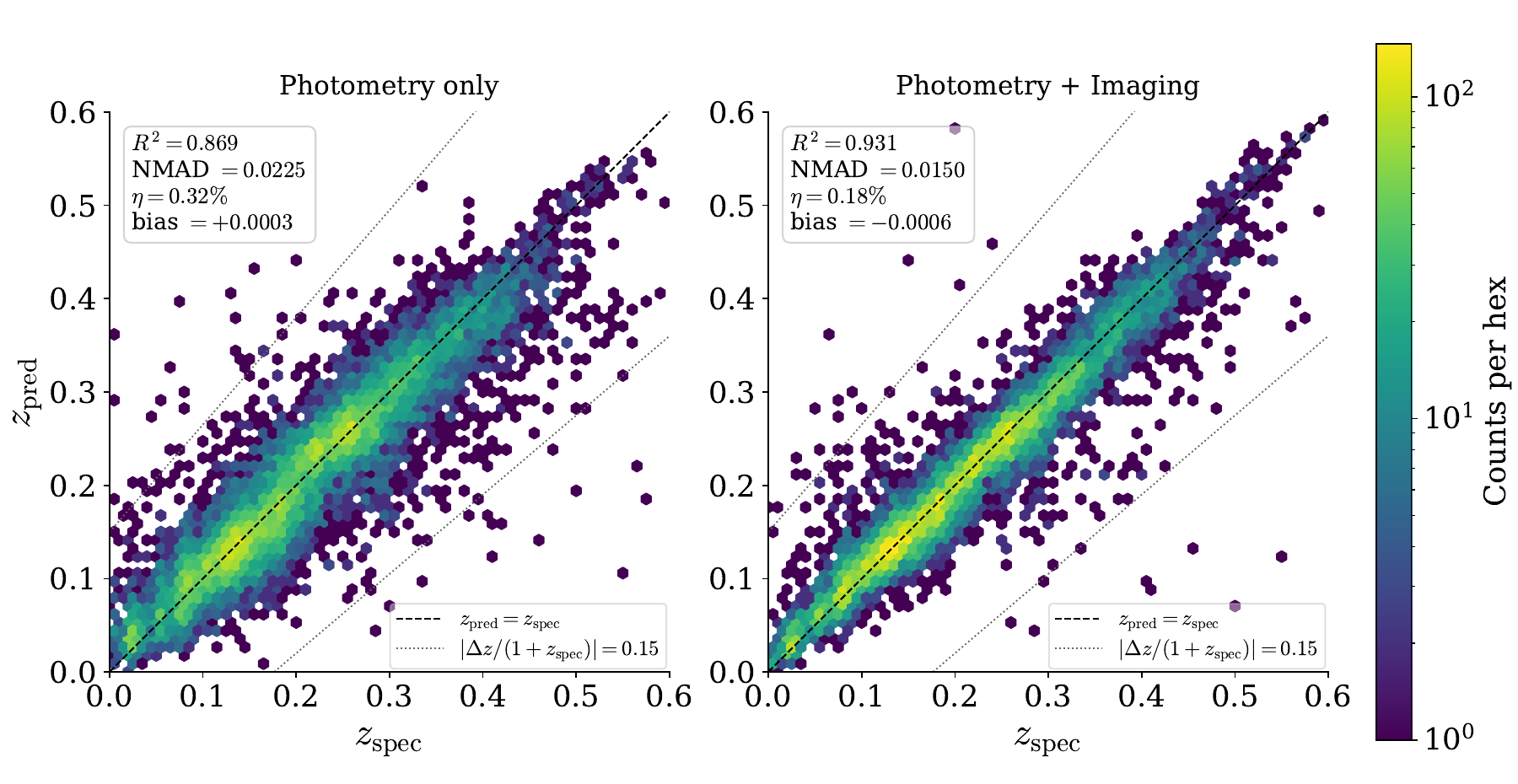}
    \caption{Predicted $z_{\rm pred}$ versus spectroscopic $z_{\rm spec}$ redshift on the test set for \emph{Ph} (left) and \emph{Ph+Im} (right) redshift readouts on frozen AION-1 encoder embeddings. Hexbin density uses a logarithmic color scale shared between panels. The dashed line marks $z_{\rm pred} = z_{\rm spec}$ and the dotted lines bound the catastrophic-outlier envelope $|\Delta z/(1+z_{\rm spec})| = 0.15$, with $\Delta z \equiv z_{\rm pred} - z_{\rm spec}$. Per-panel statistics: $R^2$; normalized median absolute deviation (NMAD), a robust scatter estimator; catastrophic-outlier fraction $\eta = $ fraction of galaxies with $|\Delta z/(1+z_{\rm spec})| > 0.15$; bias $=$ median $\Delta z/(1+z_{\rm spec})$.}
    \label{fig:fig1}
\end{figure}

\section{Results}

Fig.~\ref{fig:fig1} shows that the \emph{Ph} readout reaches $R^2 = 0.869$ and the \emph{Ph+Im} readout reaches $R^2 = 0.931$. Furthermore, adding the image modality reduces the normalized median absolute deviation from 0.0225 to 0.0150 and the catastrophic-outlier fraction from $\eta = 0.32\%$ to $0.18\%$. The median residual bias is below $10^{-3}$ in both readouts. We ask which galaxies are helped by imaging (Sec.~\ref{sec:which-galaxies}), where in the encoder the advantage enters (Sec.~\ref{sec:which-block}), and study two relevant properties (Sec.~\ref{sec:spatial-extent}--\ref{sec:dust-attenuation}).

\subsection{Imaging helps galaxies with faint fiber $r$-magnitudes and high dust attenuation}   
\label{sec:which-galaxies}

\begin{figure}[h]
    \centering
    \includegraphics[width=\textwidth]{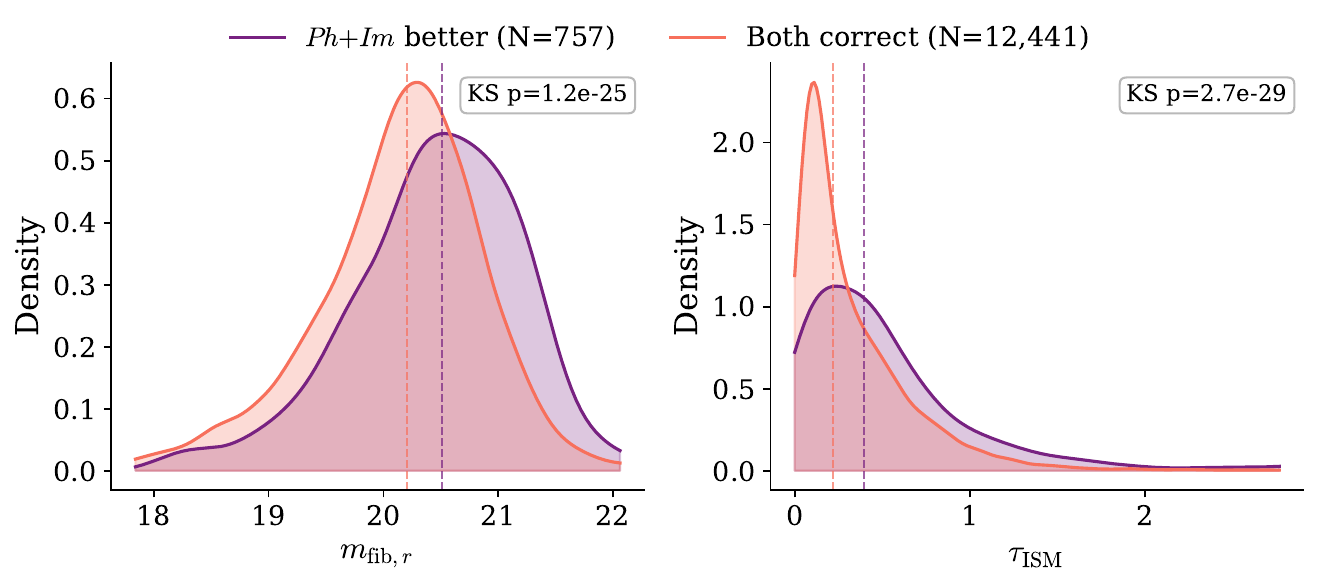}
    \caption{Kernel density estimates of two galaxy properties (left: fiber $r$-magnitude $m_{\rm fib,r}$; right: dust attenuation $\tau_{\rm ISM}$) in the held-out test set: \emph{Ph+Im better} (purple, $N=757$): galaxies for which the \emph{Ph+Im} readout satisfies $|\Delta z|/(1+z_{\rm spec}) < 0.05$ while the \emph{Ph} readout does not; \emph{Both correct} (red, $N=12{,}441$): galaxies for which both readouts satisfy this threshold. Dashed vertical lines mark per-population medians; two-sample Kolmogorov--Smirnov $p$-values are annotated in each panel.}
    \label{fig:fig2}
\end{figure}

For each test-set galaxy, we count a prediction as correct when $|\Delta z|/(1+z_{\rm spec}) < 0.05$. We partition galaxies by the joint outcome of the two readouts and focus on two non-overlapping groups: \emph{Ph+Im better} ($N=757$), galaxies that the \emph{Ph+Im} readout predicts correctly while the \emph{Ph} readout does not, and \emph{Both correct} ($N=12{,}441$), galaxies that both readouts predict correctly. Relative to the \emph{Both correct} population, \emph{Ph+Im better} galaxies have fainter fiber $r$-magnitudes $m_{\rm fib,r}$ (magnitudes are logarithmic brightness measures on which larger values are fainter) and higher dust attenuation $\tau_{\rm ISM}$ (Fig.~\ref{fig:fig2}).\footnote{Two-sample Kolmogorov--Smirnov tests reject the null of identical distributions at $p < 10^{-25}$ for both properties.} The fiber $r$-magnitude $m_{\rm fib,r}$ measures only the light falling within the $1.5''$ DESI fiber aperture, so it is fainter than the total $r$-magnitude $m_r$ by an amount that grows with the fraction of a galaxy's light lying outside the fiber. A fainter $m_{\rm fib,r}$ indicates either an intrinsically faint galaxy or a spatially extended one. Higher $\tau_{\rm ISM}$ indicates more dust in the galaxy's interstellar medium along our line of sight, which attenuates and reddens its light. Adding imaging therefore helps preferentially on galaxies that are faint within the fiber aperture and on dust-attenuated galaxies, two independent contributors to the performance gain (Appendix~\ref{app:ph_im_factors}).
  
\subsection{The multimodal advantage emerges in the first encoder block}
\label{sec:which-block}

\begin{figure}[h]
    \centering
    \includegraphics[width=\textwidth]{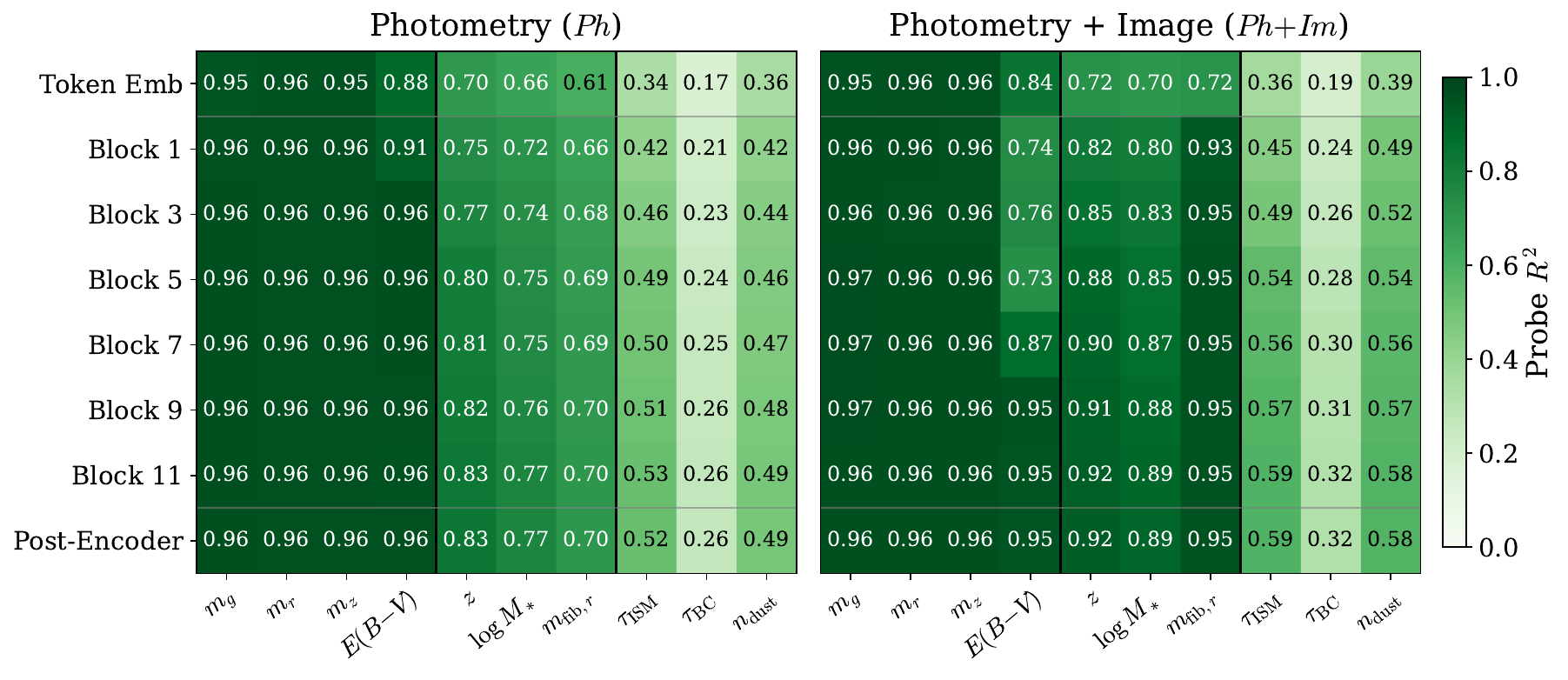}
    \caption{Layerwise linear-probe $R^2$ for 10 galaxy properties (columns) at 8 encoder stages (rows): the token embedding (``Token Emb''), the outputs of blocks 1, 3, 5, 7, 9, and 11 of the 12, and the post-encoder representation. Probes are ridge regressions on the mean-pooled 768-dim hidden state at each stage. Columns are grouped (vertical lines) into three families: broadband magnitudes ($m_g$, $m_r$, $m_z$) and $E(B{-}V)$, which are directly determined by the photometry tokens; properties whose decodability the \emph{Ph+Im} configuration substantially improves (redshift $z$, stellar mass $\log M_*$, fiber $r$-magnitude $m_{\rm fib,r}$); and dust parameters ($\tau_{\rm ISM}, \tau_{\rm BC}, n_{\rm dust}$). Left: \emph{Ph}; right: \emph{Ph+Im}.}
    \label{fig:fig3}
\end{figure}

We now ask where in the encoder the image modality's contribution enters by localizing the stage at which each galaxy property becomes linearly decodable. We fit ridge-regression linear probes to AION-1's intermediate representations at 14 stages: the token embedding (the input to block 1), the output of each of the 12 transformer blocks, and the post-encoder representation (the output of the final encoder layer-norm). We run the frozen encoder forward under both input configurations (\emph{Ph} and \emph{Ph+Im}). At each stage, we mean-pool the token activations to a single $768$-dim vector per galaxy and report the test-set $R^2$ (full procedure in Appendix~\ref{app:layerwise-probes}). 

Fig.~\ref{fig:fig3} shows the linear-probe $R^2$ for every other block and for the ten properties discussed below (all 14 stages in Fig.~\ref{fig:app-layerwise-full}). In the post-encoder representation (last row), the image modality substantially improves the decodability of three properties: redshift ($R^2 = 0.83 \to 0.92$), stellar mass ($0.77 \to 0.89$), and the fiber $r$-magnitude ($0.70 \to 0.95$). The three dust parameters shift modestly upward, but remain only weakly decodable in either configuration. Properties given directly by photometry, namely the broadband magnitudes and $E(B{-}V)$, are decodable in both configurations at the final stage ($R^2 \geq 0.95$). The image modality therefore supplies linearly decodable information about redshift, mass, fiber $r$-magnitude, and (more weakly) dust parameters.  
                  
Fig.~\ref{fig:fig3} also shows that most of the multimodal advantage is concentrated in the first encoder block. The gap between the \emph{Ph+Im} and \emph{Ph} redshift probes grows from $0.02$ at the token embedding to $0.07$ after block 1, and by only $0.02$ more over the remaining eleven blocks. Stellar mass and fiber $r$-magnitude follow a similar pattern; for the dust parameters, the probe gap is smaller and accumulates gradually across the encoder. Most of the cross-modal integration responsible for this advantage therefore occurs in block~1, motivating the analysis that follows.

We also confirm this localization with activation patching. We construct a corrupted run by replacing each galaxy's image tokens with those of another test galaxy, leaving its photometry tokens unchanged. We then patch one block's residual-stream contribution at a time, from the clean run into the corrupted run to test sufficiency, or from the corrupted run into the clean run to test necessity, and re-run the trained \emph{Ph+Im} readout (protocol and per-block values, Appendix~\ref{app:block-causal}). Block~1 is the only block that is on its own both sufficient, with its clean contribution alone restoring the readout's full $R^2 = 0.931$, and necessary, with its corrupted contribution alone driving $R^2$ to $-0.11$, close to the unpatched corrupted run's $-0.13$. 

\subsection{Spatial extent is linearly decodable and causally used, but not localized}
\label{sec:spatial-extent}
\begin{figure}[h]
    \centering
    \includegraphics[width=\textwidth]{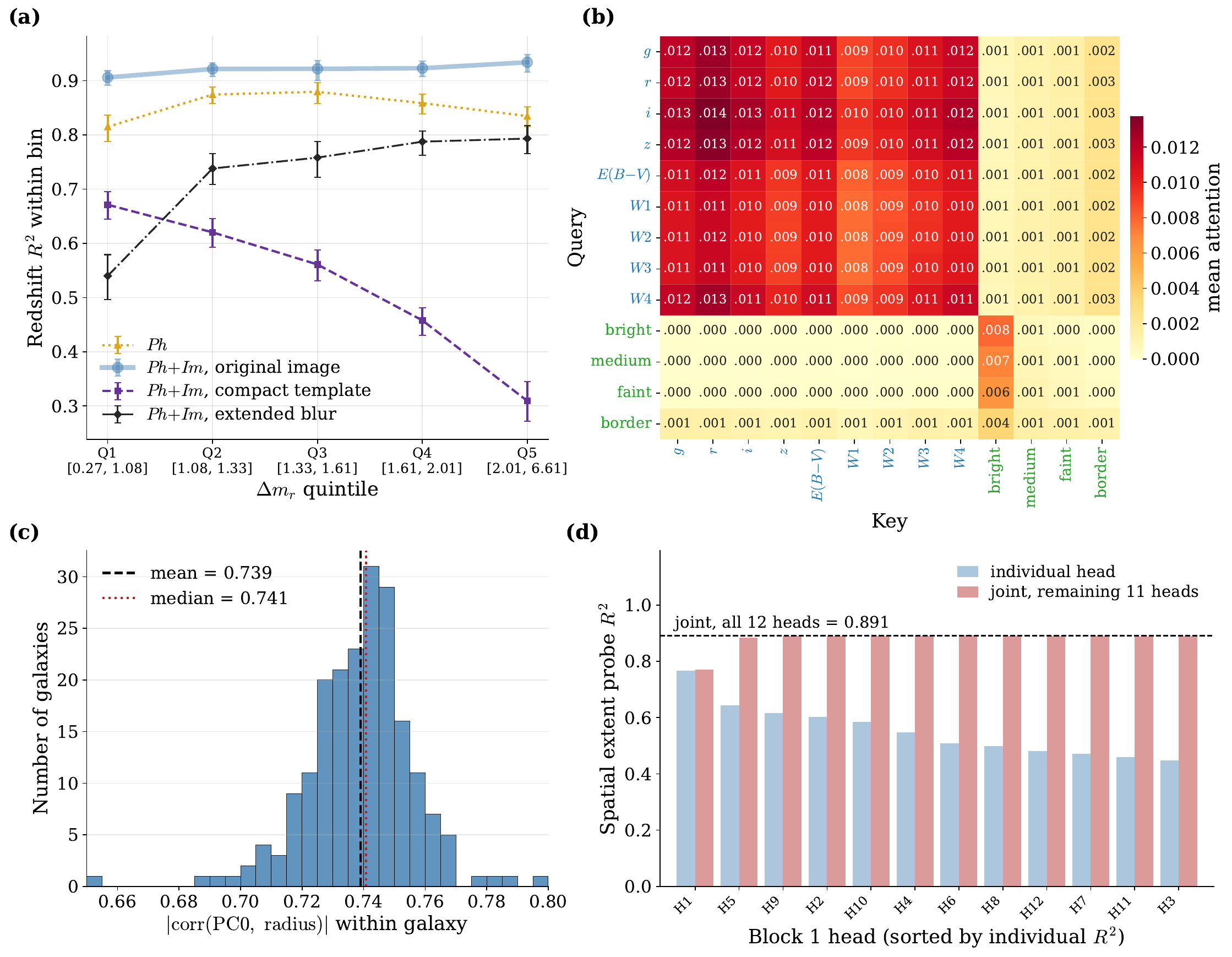}
    \caption{\textbf{(a)} Redshift $R^2$ on the held-out test set, stratified by quintiles of $\Delta m_r$, for \emph{Ph} (gold, dotted) and for \emph{Ph+Im} with the original image (blue, solid), with a compact template (purple, dashed), or with an extended blur (black, dash-dotted). Error bars are $95\%$ paired bootstrap CIs. \textbf{(b)} Mean attention computed by Head~1 under the \emph{Ph+Im} input configuration, averaged over 200 test galaxies. Keys and queries are grouped into the nine photometry tokens and four image-token regions: a two-token-wide \emph{border} ring surrounding the $24{\times}24$ token grid, and \emph{bright}/\emph{medium}/\emph{faint} quartiles (top, middle two, bottom) of token brightness within each galaxy's remaining interior. \textbf{(c)} Distribution over the same 200 galaxies of $|\mathrm{corr}(\mathrm{PC0}, \mathrm{radius})|$ across Head~1's 576 image-token outputs, where PC0 is each galaxy's first principal component. \textbf{(d)} Spatial extent probe $R^2$ from each block~1 head's $64$-dim output alone (blue) and from a probe refit on the remaining eleven heads (red), against the joint twelve-head probe (dashed, $R^2 = 0.891$).
    }
    \label{fig:fig4}
\end{figure}

The probes of Sec.~\ref{sec:which-block} show that $m_{\rm fib,r}$ information enters the representation, not that predicting redshift uses it. Moreover, $m_{\rm fib,r}$ confounds spatial extent with total brightness, since a galaxy can be faint within the $1.5''$ fiber either because it is intrinsically faint or because its light is spread beyond the aperture. For the rest of this section, we therefore stratify on the fiber-minus-total $r$-magnitude $\Delta m_r \equiv m_{\rm fib,r} - m_r = -2.5\log_{10}(F_{\rm fib}/F_{\rm tot})$, a logarithmic measure of the fraction of a galaxy's light falling outside the fiber; larger $\Delta m_r$ means a more spatially extended galaxy.
  
\subsubsection{Spatial extent is causally used}
\label{sec:extent-used}

To test whether spatial extent is causally used in predicting redshift, we first intervene on the input. For every test galaxy, we replace the four-band cutout with either a compact template (a centered two-dimensional Gaussian with $\sigma = 1.5$\,px, approximating an unresolved point source), or an extended Gaussian blur ($\sigma = 6$\,px), leaving the photometry tokens unchanged (construction of both counterfactuals, Appendix~\ref{app:image-counterfactuals}). Because the image codec reads only the central $96 \times 96$ pixels of each cutout, each counterfactual is rescaled per band so that its flux inside that crop equals the original image's; both substitutions therefore alter spatial extent without changing the brightness the codec sees. Both substantially reduce the readout's $R^2$, from the clean \emph{Ph+Im} value of $0.931$ to $0.556$ under the compact template and $0.766$ under the extended blur. Both fall below the \emph{Ph} $R^2 = 0.869$, so the readout does not revert to a photometry-only prediction when the image is replaced; the substituted morphology continues to be used.
  
Stratifying by quintiles of $\Delta m_r$ (Fig.~\ref{fig:fig4}a; per-quintile values and intervals in Table~\ref{tab:image-counterfactuals}) shows that each substitution reduces $R^2$ most where it contradicts the galaxy's true spatial extent: the compact template's decrease grows with $\Delta m_r$ ($R^2 = 0.671 \to 0.309$ from the lowest to the highest quintile), while the extended blur's shrinks ($0.540 \to 0.793$). A perturbation that merely pushes the input off-distribution would affect all quintiles alike, and the \emph{Ph} baseline varies by less than $0.07$ across the same quintiles, so neither off-distribution effects nor the bins themselves explain the trends. Redshift prediction therefore depends causally on spatial extent; Sec.~\ref{sec:extent-distributed} supports this same conclusion by erasing the spatial extent direction from block~1's output without altering any image input.
  
\subsubsection{The head that best linearly encodes spatial extent has a negligible causal effect}
\label{sec:extent-head1}

With the multimodal advantage for both redshift and $m_{\rm fib,r}$ localized to block 1, we ask which head(s) implement it. We reuse the linear-probe protocol of Sec.~\ref{sec:which-block}, applied head-wise within block~1 (Appendix~\ref{app:head1-probes}). Head~1 is the best of the 12 block 1 heads on seven of the eight probe targets (Table~\ref{tab:head-table}), including spatial extent ($\Delta m_r$: $R^2 = 0.767$ against $0.643$ of the next best head) and redshift ($z$: $0.697$ against $0.594$). The only exception is $\tau_{\rm ISM}$.

Head~1 is consistent with a head that computes spatial extent. In the attention pattern (Fig.~\ref{fig:fig4}b), image queries in the galaxy interior (\emph{bright}, \emph{medium}, and \emph{faint}) attend most strongly to \emph{bright} image keys. This image-token aggregation is organized radially: the first principal component (PC0) of Head~1's \emph{Ph+Im} output explains a mean $84.1\%$ of the variance among a galaxy's 576 image tokens, and each token's PC0 score tracks its distance from the image center, with $|\mathrm{corr}(\mathrm{PC0}, \mathrm{radius})|$ of mean $0.739$ (Fig.~\ref{fig:fig4}c; Appendix~\ref{app:head1-probes}). A dominant output direction that tracks radial position is what a head encoding spatial extent would produce.

However, zeroing Head~1's output for the test set and running the trained \emph{Ph+Im} redshift readout moves $R^2$ by $< 0.002$ in any $\Delta m_r$ quintile (Appendix~\ref{app:head1-probes}). Replacing Head~1's output at all $585$ positions with that of a redshift-matched galaxy from the opposite $\Delta m_r$ quintile, an in-distribution intervention, likewise has a negligible effect on the prediction (Appendix~\ref{app:head1-patching}). The head that best linearly encodes spatial extent therefore has a negligible causal effect on the prediction.
  
\subsubsection{Spatial extent is not localized to a single head or direction}
\label{sec:extent-distributed}

The results of Sec.~\ref{sec:extent-used} and \ref{sec:extent-head1} fit together because spatial extent is not localized to Head~1. Head~1 is the single best head for encoding $\Delta m_r$ ($R^2 = 0.767$), but a probe refit on the eleven heads that remain when Head~1 is removed reaches $R^2 = 0.771$ (Fig.~\ref{fig:fig4}d; joint and leave-one-head-out probes, Appendix~\ref{app:head1-probes}): spatial extent remains decodable from block~1 after Head~1 is ablated. Additionally, every other head in block~1 carries spatial extent information on its own ($R^2$ between $0.448$ and $0.643$, Table~\ref{tab:head-table}), a probe on all twelve jointly reaches $0.891$, and removing any one of the other eleven heads from the joint probe reduces its $R^2$ by at most $0.006$. Therefore, no single head is necessary for spatial extent to be linearly decodable from block~1's head outputs.
  
Erasing the linear spatial extent direction from block~1's full output, rather than ablating a head, does not remove the property either. A rank-1 linear concept eraser (LEACE; \citealp{belrose2023leace}) makes $\Delta m_r$ linearly undecodable from block~1's mean-pooled output ($R^2 = 0.903 \to 0$), but it is decodable again from the residual stream immediately after block~2's attention ($R^2 = 0.909$, against a clean $0.912$). The erasure does reduce redshift decodability downstream, however: with the linear spatial extent direction removed at block~1, the post-encoder redshift probe falls from $R^2 = 0.918$ to $0.865$, a loss of $0.054$ against $0.011$ for a shuffled-label control eraser, so the spatial extent representation that remains decodable downstream is not a complete substitute for the erased one (Appendix~\ref{app:extent-erasure}). Spatial extent is therefore used but not localized: no single head is necessary for it, and erasing its linear direction in block~1 does not remove it from the representation.

\subsection{Dust attenuation is weakly decodable, yet the dust-graded advantage localizes to one head}
\label{sec:dust-attenuation}

\begin{figure}[h]
      \centering                                      \includegraphics[width=\textwidth]{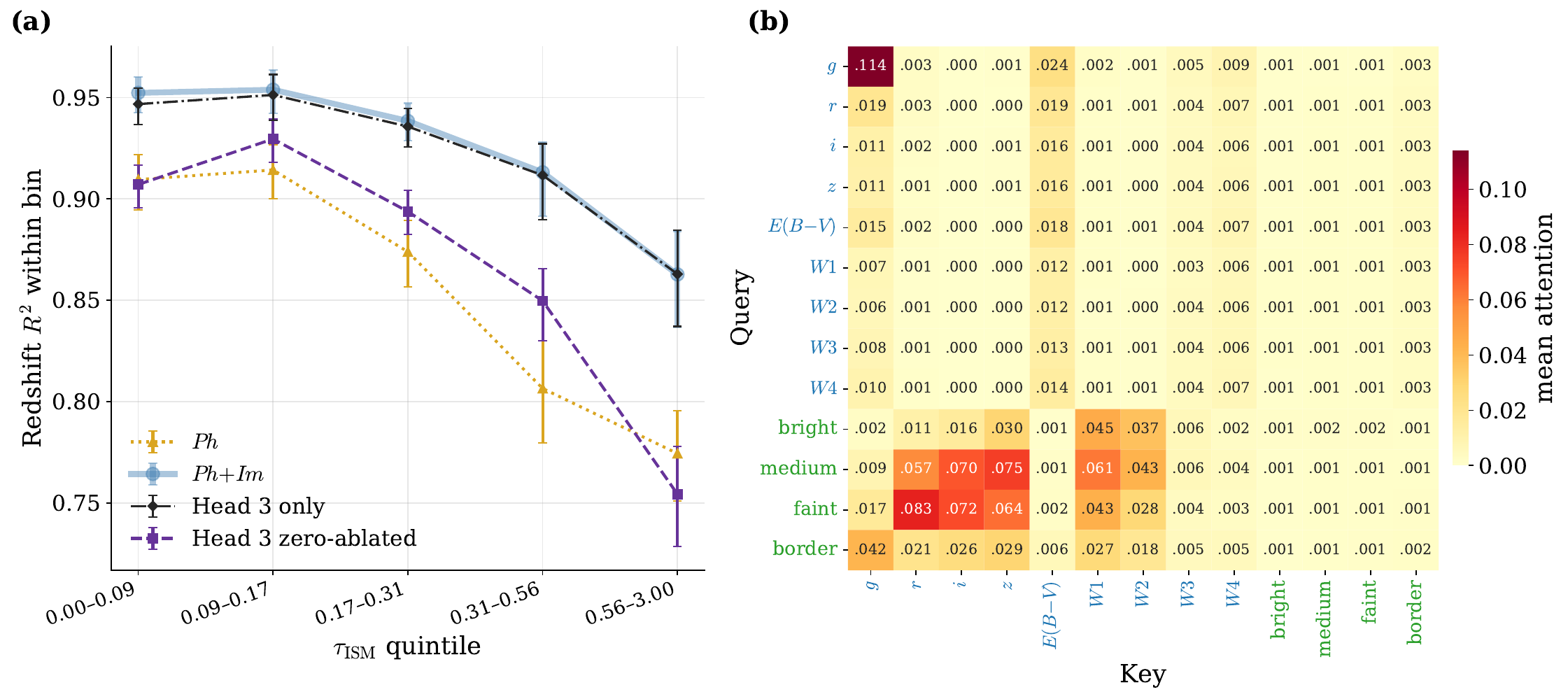}
      \caption{\textbf{(a)} Redshift $R^2$ on the held-out test set, stratified by quintiles of $\tau_{\rm ISM}$, for \emph{Ph} (gold, dotted), \emph{Ph+Im} (blue, solid), \emph{Ph+Im} with the eleven other block-1 heads mean-ablated so Head~3 is the only active head (black, dash-dotted), and \emph{Ph+Im} with Head~3's output zeroed (purple, dashed). Mean ablation replaces a head's output with its per-position mean over a random subsample of $5{,}000$ training-split galaxies. Error bars are $95\%$ paired bootstrap CIs. \textbf{(b)} Mean \emph{Ph+Im} attention from Head~3, averaged over 200 held-out test galaxies, grouped as in Fig.~\ref{fig:fig4}b.}                                 
      \label{fig:fig5}    
  \end{figure}     
  
\subsubsection{Head 3 is necessary for the dust-graded multimodal advantage}

\label{sec:head3-necessary}

Sec.~\ref{sec:extent-head1} established that the head with the best-performing linear probes in block~1 has a negligible causal effect.  We therefore searched block~1 by ranking all twelve heads by causal effect rather than by linear decodability: we zero-ablated each head's output at every position and re-ran the trained \emph{Ph+Im} redshift readout on the whole test set and within the most dust-attenuated $\tau_{\rm ISM}$ quintile (Table~\ref{tab:head-table}). Zero-ablating Head~3 gives $\Delta R^2 = -0.052$ overall, against $-0.003$ for the next-largest head in the block. Head~3's rank is robust to multiplicity correction, retraining the readout on different early-stopping criteria and data splits, and the choice of ablation (Appendix~\ref{app:head3-robustness}). The properties on which we stratify were fixed by the population analysis of Sec.~\ref{sec:which-galaxies} before any head was examined, so the stratified test is not a post-hoc search over subgroups.

From the least to the most dust-attenuated quintile, ablating Head~3 reduces $R^2$ in every quintile, with the largest effect in the most dust-attenuated one ($\Delta R^2 = -0.108$; all values and intervals in Table~\ref{tab:h3-quintiles}). This is where the multimodal advantage is concentrated: $R^2_{\rm Ph+Im} - R^2_{\rm Ph}$ is about twice as large on the two most dust-attenuated quintiles ($\approx 0.10$) as on the two least ($\approx 0.04$), with non-overlapping CIs, and ablating Head~3 removes most of it in every quintile (residual advantage between $-0.02$ and $+0.04$). In the most dust-attenuated quintile, removing this single head reduces the \emph{Ph+Im} readout's $R^2$ from $0.863$ to $0.754$, close to the \emph{Ph} value of $0.774$.
  
Head~3 is also sufficient within block~1: when the other eleven block-1 heads are replaced by their per-position mean, so that Head~3 is the only block-1 head whose output depends on the input galaxy, the readout reaches $R^2 = 0.928$ against $0.931$ for the \emph{Ph+Im} readout, and the gap is smallest in the quintiles where the ablation effect is largest: Head~3 alone recovers at least $87\%$ of the multimodal advantage in every quintile and $100\%$ in the most dust-attenuated one (Fig.~\ref{fig:fig5}a; mean-ablation protocol and per-quintile values, Appendix~\ref{app:head3-robustness}). Head~3 is therefore necessary for most of the advantage in every dust stratum and, among the block-1 heads, sufficient for nearly all of it.

Ablating Head~3 also changes the kind of error the model makes. We fit $z_{\rm pred} = \alpha + \beta z_{\rm spec}$ and call a slope $\beta < 1$ \emph{attenuation bias}: a statistical term for predictions compressed toward the mean redshift. \citet{moran2026regions} find that image-based photometric-redshift methods suffer less of it than photometry-based ones. Consistent with this, adding imaging reduces attenuation bias in the most dust-attenuated quintile, raising $\beta$ from $0.806$ (\emph{Ph}) to $0.867$ (\emph{Ph+Im}). Ablating Head~3 does not return the slope to its \emph{Ph} value but lowers it further, to $0.710$, while the scatter about the fitted line stays slightly below that of \emph{Ph} ($0.039$ against $0.042$; decomposition in Appendix~\ref{app:slope-bias}). In this quintile, the reduction in attenuation bias that imaging provides therefore requires Head~3: with Head~3 ablated, \emph{Ph+Im} is more attenuation-biased than  \emph{Ph}.
  
\subsubsection{Head 3's computation remains unidentified}
\label{sec:head3-negative}

Head~3 performs cross-modal attention: its image queries place a mean of $24.9\%$ of their attention on the nine photometry keys, concentrating on the $r$, $i$, $z$  flux tokens and the $W1$ and $W2$ infrared flux tokens (Fig.~\ref{fig:fig5}b). We tested four families of candidate operations, drawn from that attention pattern and from the standard photometric-redshift toolbox, and found support for none of them. Protocols and full results are given in Appendices~\ref{app:head3-probes}--\ref{app:head3-cross-modal}.

\begin{enumerate}
    \item \textbf{Probes for expected physical parameters.} If Head~3 encoded the physical parameters that make dust-attenuated galaxies hard to predict, probes on its output would recover them. It ranks in the middle of the twelve heads for redshift, last for $\Delta m_r$, and first but weakly for $\tau_{\rm ISM}$ ($R^2 = 0.264$ linear, $0.397$ MLP, against $0.589$ at the post-encoder). Its dust-graded causal effect is not mirrored by decodability (Appendix~\ref{app:head3-probes}).
    
    \item \textbf{Dereddening and break tracking.} Among the photometry keys, every photometry query attends most to the $g$ flux token and $E(B{-}V)$ (Fig.~\ref{fig:fig5}b), suggesting dereddening, which corrects fluxes for Milky Way dust, or tracking of the $4000\,$\AA\ break, whose redshift-dependent position is the classic photometric-redshift signal. Neither is supported: Head~3's output at the $g$ flux position does not vary with $E(B{-}V)$ (Pearson correlation $-0.03$), its attention does not shift from $g$ to $r$ with redshift, and it ranks seventh of twelve heads on the redshift-corrected $g{-}r$ color, the break's photometric signature (Appendix~\ref{app:head3-photometry-attn}).
    
    \item \textbf{Image-token spatial encoding.} If Head~3 encoded resolved spatial structure, its image-token output would track within-galaxy color or brightness maps and its ablation would matter most for galaxies with spatially variable color. Neither is supported (mean within-galaxy $|\mathrm{corr}| \le 0.13$; the ablation effect is smaller for such galaxies), and no spatial subset of image tokens carries more than $41.9\%$ of the effect. Head~3's effect is distributed across the image tokens rather than localized (Appendix~\ref{app:head3-image}).
    
    \item \textbf{Correction of photometry--image discrepancies.} If Head~3's effect came from correcting inconsistencies between the photometry-token and image-token representations, it would grow with $|D|$, the discrepancy between the $g$-band magnitudes decoded from the two. Instead, the $R^2$ loss from ablating Head~3 shrinks, from $0.066$ in the lowest-$|D|$ quintile to $0.031$ in the highest, and $|D|$ is uncorrelated with $\tau_{\rm ISM}$ (Appendix~\ref{app:head3-cross-modal}).

\end{enumerate}

Because each test presupposes a candidate operation, we also asked whether Head~3's output contains the redshift information that distinguishes \emph{Ph+Im} from \emph{Ph}, regardless of what it encodes: a redshift probe on the \emph{Ph} representation concatenated with Head~3's pooled output closes only $5\%$ of the gap between the \emph{Ph} and \emph{Ph+Im} redshift probes, and a token-resolved probe over the $576$ image tokens closes only $10\%$ (Appendix~\ref{app:head3-concat}). The information that distinguishes the two representations therefore cannot be extracted from Head~3's output by any pooled or token-resolved probe we fit.

\subsubsection{No comparable head or dominant relay among the heads tested accounts for the effect}
\label{sec:head3-exhaustive}

Since the search was restricted to the twelve heads of block~1, two alternatives remain open: that the computation is shared with heads elsewhere in the encoder, or that Head~3's output is merely the entry point of a circuit carried out downstream. We screened all $144$ encoder heads with a gradient$\times$activation attribution and zero-ablated the five heads outside block~1 that rank highest on the most dust-attenuated galaxies, four of which exceed our screening threshold in that quintile; we call these the confirmed heads (Appendix~\ref{app:head3-circuit}). All five have effects that are statistically significant but small over the full test set, the largest being $\Delta R^2 = -0.005$ against Head~3's $-0.052$.

We then tested whether the four confirmed heads mediate Head~3's effect, using two interventions (Appendix~\ref{app:head3-circuit}). Single-edge path patching recomputes one receiving head's queries, keys, and values from a residual stream with Head~3's output removed, while every other head reads the clean residual stream. Cutting the edge into any one confirmed head accounts for at most $7\%$ of Head~3's $\Delta R^2$, against at most $0.2\%$ for control heads. Because path patching cuts one edge at a time, we also zero-ablate Head~3 while fixing all four confirmed heads to their clean-run outputs, so that the ablation cannot propagate through them. In the most dust-attenuated quintile, freezing them removes only $3\%$ of the ablation effect, comparable in magnitude to freezing four control heads. No dominant relay therefore runs through these heads. Because the search covered attention heads only, contributions through the MLPs, through many weak edges jointly, or directly through the residual stream are not excluded, and the mechanism remains unidentified.

\section{Discussion}

To our knowledge, this is the first study to apply causal mechanistic interpretability methods to an astronomy foundation model. The multimodal advantage over photometry alone is concentrated in spatially extended and dust-attenuated galaxies and mostly enters in AION-1's first encoder block, where linear decodability and causal importance dissociate in two opposite ways. Spatial extent is causally used but not localized: the head that decodes it best has a negligible causal effect, and the property remains decodable after ablation of any single head and after erasure of its linear direction. A strong probe therefore does not identify the causally relevant component, a null ablation does not establish that a property is unused, and methods that assume a property is carried by a single head or linear direction have no such target here. Interpreting scientific foundation models will require methods that characterize distributed representations of this kind rather than only localize them. 
  
Dust attenuation is the opposite case: a single head is necessary for most of the dust-graded advantage and, among its block's heads, sufficient for nearly all of it, yet the dust parameters are only weakly decodable from its output, and none of the astrophysically motivated operations we tested predicts its behavior. Here the computation is localized, and the open problem is characterizing the function the head implements. Because our circuit search covered attention heads only and our hypothesis tests are limited to the operations we could enumerate, we cannot tell whether the head computes something that matches no known astrophysically motivated operation or something our methods cannot recover. Distinguishing them requires a method that extracts the function a component computes in a form a domain scientist can evaluate, rather than testing candidate operations one at a time. 

Connecting model internals to domain knowledge is, we believe, a prerequisite for extracting scientific hypotheses from these models, and astronomy is a useful setting in which to build it because the domain supplies a ground truth most interpretability settings lack: whether a model implements the operation an astrophysicist would expect has a determinate answer, and either answer is informative.

\paragraph{Scope and limitations.} The single-model, single-task focus of our paper is deliberate. Mechanistic interpretability of scientific foundation models is, we argue, most tractable when anchored to a specific downstream task: the task provides an output whose change measures each component's causal effect, and the domain hypotheses to test against. The task then fixes the remaining choices: redshift prediction requires galaxies with imaging, photometry, and spectroscopy, so we use the largest such sample among AION-1's training surveys, which may affect absolute accuracy but not our claims about how the frozen encoder computes the gain. Comparison across astronomical foundation models is a natural extension but not an architecturally direct one, since contrastive models such as AstroCLIP and AstroM$^3$ couple modalities only through the training objective. Our results are therefore for one model and dataset; how they change with scale or survey is left for future work.
 
\begin{ack}
R.C. Zhang is supported by a Gates Cambridge Scholarship from the Gates Cambridge Trust. A. Favero acknowledges support from the Infosys-Cambridge AI Centre. M. Cranmer is grateful for support from the Schmidt Sciences AI2050 Early Career Fellowship and the Isaac Newton Trust. Polymathic AI acknowledges support provided by the Simons Foundation and Schmidt Sciences, LLC.

\paragraph{Data.} This research used data obtained with the Dark Energy Spectroscopic Instrument (DESI). DESI construction and operations is managed by the Lawrence Berkeley National Laboratory. This material is based upon work supported by the U.S. Department of Energy, Office of Science, Office of High-Energy Physics, under Contract No. DE--AC02--05CH11231, and by the National Energy Research Scientific Computing Center, a DOE Office of Science User Facility under the same contract. Additional support for DESI was provided by the U.S. National Science Foundation (NSF), Division of Astronomical Sciences under Contract No. AST-0950945 to the NSF's National Optical-Infrared Astronomy Research Laboratory; the Science and Technology Facilities Council of the United Kingdom; the Gordon and Betty Moore Foundation; the Heising-Simons Foundation; the French Alternative Energies and Atomic Energy Commission (CEA); the National Council of Humanities, Science and Technology of Mexico (CONAHCYT); the Ministry of Science and Innovation of Spain (MICINN), and by the DESI Member Institutions: \url{www.desi.lbl.gov/collaborating-institutions}. The DESI collaboration is honored to be permitted to conduct scientific research on I'oligam Du'ag (Kitt Peak), a mountain with particular significance to the Tohono O'odham Nation. Any opinions, findings, and conclusions or recommendations expressed in this material are those of the author(s) and do not necessarily reflect the views of the U.S. National Science Foundation, the U.S. Department of Energy, or any of the listed funding agencies.

The Legacy Surveys consist of three individual and complementary projects: the Dark Energy Camera Legacy Survey (DECaLS; Proposal ID \#2014B-0404; PIs: David Schlegel and Arjun Dey), the Beijing-Arizona Sky Survey (BASS; NOAO Prop. ID \#2015A-0801; PIs: Zhou Xu and Xiaohui Fan), and the Mayall z-band Legacy Survey (MzLS; Prop. ID \#2016A-0453; PI: Arjun Dey). DECaLS, BASS and MzLS together include data obtained, respectively, at the Blanco telescope, Cerro Tololo Inter-American Observatory, NSF's NOIRLab; the Bok telescope, Steward Observatory, University of Arizona; and the Mayall telescope, Kitt Peak National Observatory, NOIRLab. Pipeline processing and analyses of the data were supported by NOIRLab and the Lawrence Berkeley National Laboratory (LBNL). The Legacy Surveys project is honored to be permitted to conduct astronomical research on I'oligam Du'ag (Kitt Peak), a mountain with particular significance to the Tohono O'odham Nation. NOIRLab is operated by the Association of Universities for Research in Astronomy (AURA) under a cooperative agreement with the National Science Foundation. LBNL is managed by the Regents of the University of California under contract to the U.S. Department of Energy. This project used data obtained with the Dark Energy Camera (DECam), which was constructed by the Dark Energy Survey (DES) collaboration. Funding for the DES Projects has been provided by the U.S. Department of Energy, the U.S. National Science Foundation, the Ministry of Science and Education of Spain, the Science and Technology Facilities Council of the United Kingdom, the Higher Education Funding Council for England, the National Center for Supercomputing Applications at the University of Illinois at Urbana-Champaign, the Kavli Institute of Cosmological Physics at the University of Chicago, Center for Cosmology and Astro-Particle Physics at the Ohio State University, the Mitchell Institute for Fundamental Physics and Astronomy at Texas A\&M University, Financiadora de Estudos e Projetos, Fundacao Carlos Chagas Filho de Amparo, Financiadora de Estudos e Projetos, Fundacao Carlos Chagas Filho de Amparo a Pesquisa do Estado do Rio de Janeiro, Conselho Nacional de Desenvolvimento Cientifico e Tecnologico and the Ministerio da Ciencia, Tecnologia e Inovacao, the Deutsche Forschungsgemeinschaft and the Collaborating Institutions in the Dark Energy Survey. The Collaborating Institutions are Argonne National Laboratory, the University of California at Santa Cruz, the University of Cambridge, Centro de Investigaciones Energeticas, Medioambientales y Tecnologicas-Madrid, the University of Chicago, University College London, the DES-Brazil Consortium, the University of Edinburgh, the Eidgenossische Technische Hochschule (ETH) Zurich, Fermi National Accelerator Laboratory, the University of Illinois at Urbana-Champaign, the Institut de Ciencies de l'Espai (IEEC/CSIC), the Institut de Fisica d'Altes Energies, Lawrence Berkeley National Laboratory, the Ludwig Maximilians Universitat Munchen and the associated Excellence Cluster Universe, the University of Michigan, NSF's NOIRLab, the University of Nottingham, the Ohio State University, the University of Pennsylvania, the University of Portsmouth, SLAC National Accelerator Laboratory, Stanford University, the University of Sussex, and Texas A\&M University. BASS is a key project of the Telescope Access Program (TAP), which has been funded by the National Astronomical Observatories of China, the Chinese Academy of Sciences (the Strategic Priority Research Program ``The Emergence of Cosmological Structures'' Grant \# XDB09000000), and the Special Fund for Astronomy from the Ministry of Finance. The BASS is also supported by the External Cooperation Program of Chinese Academy of Sciences (Grant \# 114A11KYSB20160057), and Chinese National Natural Science Foundation (Grant \# 12120101003, \# 11433005). The Legacy Survey team makes use of data products from the Near-Earth Object Wide-field Infrared Survey Explorer (NEOWISE), which is a project of the Jet Propulsion Laboratory/California Institute of Technology. NEOWISE is funded by the National Aeronautics and Space Administration. The Legacy Surveys imaging of the DESI footprint is supported by the Director, Office of Science, Office of High Energy Physics of the U.S. Department of Energy under Contract No. DE-AC02-05CH11231, by the National Energy Research Scientific Computing Center, a DOE Office of Science User Facility under the same contract; and by the U.S. National Science Foundation, Division of Astronomical Sciences under Contract No. AST-0950945 to NOAO.

\end{ack}

\bibliographystyle{plainnat}
\bibliography{neurips_2026}

\medskip


\appendix
\counterwithin{figure}{section}
\counterwithin{table}{section}
  
\section{Readout architecture, training, data split, and confidence intervals}
\label{app:redshift_pred}
\paragraph{Architecture.} The readout is a cross-attention pooling layer followed by a two-layer MLP, trained on the output of the frozen AION-1 encoder. The pooling layer is a \texttt{NormCrossAttention} block (12 heads, $d_{\text{model}} = 768$, no QKV or projection bias, no attention dropout, per-head query-key normalization) whose single learnable query of dimension $768$ aggregates the encoder tokens into a single $768$-dimensional pooled vector. The MLP maps $768 \to 256 \to 1$ with a GELU activation and produces a scalar redshift estimate. Under \emph{Ph}, the encoder receives only the nine photometry tokens and the readout pools over the nine corresponding encoder outputs; under \emph{Ph+Im} the encoder additionally receives the $576$ image tokens produced by the image codec, and the readout pools over all $585$. Targets are PROVABGS spectroscopic redshifts (\texttt{provabgs\_desi\_edr\_sv3/Z\_HP}) standardized by the training-split mean and standard deviation, and the training loss is the mean-squared error on the standardized target.

\paragraph{Inputs.} Photometry inputs are the nine Legacy Survey scalar tokens used during AION-1 pretraining: the $g$, $r$, $i$, $z$, $W1$, $W2$, $W3$, and $W4$ fluxes and the foreground extinction $E(B{-}V)$. Image inputs are four-band Legacy Survey cutouts of size $160 \times 160$ pixels, passed to AION-1 with the band labels \texttt{DES-G}, \texttt{DES-R}, \texttt{DES-I}, and \texttt{DES-Z}. AION-1's image codec first center-crops each cutout to $96 \times 96$ pixels, clamps pixel values to a fixed range, rescales them to the Legacy Survey flux convention, applies an $\mathrm{arcsinh}$ range compression (compression factor $0.01$, post-multiplication factor $10$), and pads the four input bands into AION-1's multi-survey band-slot layout. The result is passed through the \texttt{MagVitAE} convolutional autoencoder (\texttt{hidden\_dims}$\,=\,$512, \texttt{n\_compressions}$\,=\,$2, \texttt{num\_consecutive}$\,=\,$4, finite-scalar quantization), whose two consecutive $2{\times}$ spatial downsampling stages yield a $24 \times 24 = 576$-token map per galaxy.

\paragraph{Training.} Both the \emph{Ph} and \emph{Ph+Im} readouts are trained from random initialization for up to $200$ epochs with AdamW (learning rate $5 \times 10^{-4}$, weight decay $0.05$) and batch size $32$. The learning-rate schedule is a $5$-epoch linear warmup from $1\%$ of the peak learning rate up to $5 \times 10^{-4}$, followed by cosine annealing over the remaining $195$ epochs to a floor of $10^{-6}$. Gradient norms are clipped to $\|g\|_2 \le 1.0$. Training images are augmented with random horizontal and vertical flips at every batch, and no augmentation is applied at evaluation. Optimization is performed only on the cross-attention-pooling and MLP parameters; the AION-1 encoder and modality codecs are kept in evaluation mode with \texttt{requires\_grad}$\,=\,$\texttt{False} and produce no gradients. We early-stop on the held-out test-set $R^2$ with a patience of $15$ epochs without improvement and restore the best-$R^2$ checkpoint for all subsequent analyses. Because early stopping on the test set uses the same galaxies we report results on, we also retrained both readouts with early stopping on a validation set drawn from the training split ($20\%$ of the training split, \texttt{random\_state}$\,=\,$\texttt{7}), once on the seed-42 partition used throughout the paper and once on a second partition drawn with seed 123. The results are stable. On the seed-42 partition, the retrained readouts reach test-set $R^2 = 0.870\,[0.863, 0.877]$ for \emph{Ph} and $0.930\,[0.924, 0.936]$ for \emph{Ph+Im}, against $0.869\,[0.861, 0.877]$ and $0.931\,[0.924, 0.936]$ for the originals; on the seed-123 partition, they reach $0.867\,[0.859, 0.875]$ and $0.927\,[0.920, 0.934]$; and the twelve-head ablation ranking of Sec.~\ref{sec:head3-necessary} is reproduced on both partitions (Appendix~\ref{app:head3-robustness}). We therefore use the readouts trained with test-set early stopping throughout the paper.

\paragraph{Data split.} The training and held-out splits used throughout the paper consist of $54{,}204$ and $13{,}551$ galaxies respectively, obtained by an $80{/}20$ random partition (\texttt{sklearn.model\_selection.train\_test\_split}, \texttt{random\_state}$\,=\,$\texttt{42}) of the $67{,}755$ DESI EDR BGS-bright galaxies that pass the quality cuts of Sec.~\ref{sec:setup}, namely \texttt{IS\_BGS\_BRIGHT}, $z_{\rm spec} > 0$, $\log M_\star > 0$, positive PROVABGS optical and $W1$ magnitudes, and finite Legacy Survey $g/r/i/z$ fluxes. 

\paragraph{Bootstrap confidence intervals.} 
\label{app:bootstrap}
Unless stated otherwise, every bracketed interval and error bar in this paper is a $95\%$ percentile bootstrap interval, the $2.5$\textsuperscript{th}--$97.5$\textsuperscript{th} percentile range of the bootstrap distribution, computed from $B = 2{,}000$ resamples. For readout $R^2$ values and their differences, whether between two readouts or between a readout with and without an intervention, each resample draws the relevant galaxies (the whole test set, or a subset of it such as one $\tau_{\rm ISM}$ quintile) with replacement, and every readout and intervention being compared is evaluated on that same resampled set. This pairing matters because galaxies that are hard for one readout tend to be hard for the others, so their per-galaxy errors are positively correlated, and resampling each one separately would overstate the uncertainty on their differences. The correlation is strongest between a readout and an intervention on that same readout, which is why an interval on $\Delta R^2$ is narrower than the intervals on the two $R^2$ values it is formed from. Intervals on probe $R^2$ are computed for each probe separately, with the probe held fixed and its held-out galaxies resampled. All intervals reflect only sampling variation across galaxies, not variation from retraining a readout or probe.

\section{Independent contributions of $m_{\rm fib,r}$ and $\tau_{\rm ISM}$ to the multimodal advantage}              
\label{app:ph_im_factors}
  
The fiber magnitude $m_{\rm fib,r}$ and the dust attenuation $\tau_{\rm ISM}$ could in principle reflect a single underlying mechanism, since dust attenuation dims a galaxy and would raise $m_{\rm fib,r}$ along with $\tau_{\rm ISM}$. Three statistical analyses indicate that they make independent contributions to the multimodal accuracy gain. All three treat a galaxy as ``correct'' under each model when $|\Delta z|/(1+z_{\rm spec}) < 0.05$. 

\paragraph{Marginal correlation.} On the $13{,}198$ galaxies in the \emph{Ph+Im better} and \emph{Both correct} populations of Sec.~\ref{sec:which-galaxies} combined, the Spearman rank correlation between $m_{\rm fib,r}$ and $\tau_{\rm ISM}$ is weak, $\rho = 0.15$ (\texttt{scipy.stats.spearmanr}). 

\paragraph{Joint logistic regression.} We ask whether each of $m_{\rm fib,r}$ and $\tau_{\rm ISM}$ predicts whether a galaxy is \emph{Ph+Im better} after controlling for the other and for three further properties. The sample is the same $13{,}198$ galaxies, with binary label $y = 1$ for \emph{Ph+Im better} and $y = 0$ for \emph{Both correct}. The feature vector $\mathbf{x} \in \mathbb{R}^5$ consists of $m_{\rm fib,r}$, $\tau_{\rm ISM}$, $n_{\rm dust}$, $\log M_\star$, and $m_r$, each standardized to zero mean and unit variance over the sample. We fit
  \begin{equation}
  P(y = 1 \mid \mathbf{x}) = \sigma\!\left(\beta_0 + \boldsymbol{\beta}^{\top}\mathbf{x}\right), \qquad \sigma(u) = 1/(1 + e^{-u}),
  \end{equation}
by maximum likelihood without regularization (\texttt{sklearn.linear\_model.LogisticRegression} with \texttt{C}$\,=\,$\texttt{np.inf}, L-BFGS solver). Standard errors (SE) are the square roots of the diagonal of the inverse observed Fisher information $(X^{\top} W X)^{-1}$, where $X$ is the $N \times 6$ design matrix with an intercept column, $W = \mathrm{diag}\big(\hat p_i(1 - \hat p_i)\big)$, and $\hat p_i$ is the fitted probability for galaxy $i$; $95\%$ Wald CIs are $\hat\beta_j \pm 1.96\,\mathrm{SE}_j$. Because the features are standardized, $\hat\beta_j$ is the change in log-odds of being \emph{Ph+Im better} per one standard deviation of feature $j$ with the other four held fixed. Both coefficients of interest are positive with CIs excluding zero: $\hat\beta_{m_{\rm fib,r}} = +0.62 \pm 0.11$ and $\hat\beta_{\tau_{\rm ISM}} = +0.28 \pm 0.07$. The five-feature model reaches an in-sample AUC of $0.666$. Each property therefore predicts whether a galaxy is \emph{Ph+Im better} independently of the other and of the three controls.
  
\paragraph{Interaction test.} On the full held-out test set ($N = 13{,}551$; both $m_{\rm fib,r}$ and $\tau_{\rm ISM}$ are available for every test galaxy, so no galaxies are dropped), we split galaxies at the medians of the two properties, $m_{\rm fib,r} = 20.23$ and $\tau_{\rm ISM} = 0.23$, and compute the accuracy gain $\Delta = p_1 - p_2$ in each of the four cells, where $p_1$ and $p_2$ are the fractions of the cell's galaxies predicted correctly by the \emph{Ph+Im} and \emph{Ph} readouts, respectively (Table~\ref{tab:2x2}). The gain increases along both axes: $+2.2\%$ in the (low $m_{\rm fib,r}$, low $\tau_{\rm ISM}$) cell, $+4.7\%$ when $m_{\rm fib,r}$ alone is high, $+5.1\%$ when $\tau_{\rm ISM}$ alone is high, and $+7.8\%$ when both are high. To test whether the two properties interact, we define the indicators $a = \mathbf{1}[m_{\rm fib,r} \ge 20.23]$ and $b = \mathbf{1}[\tau_{\rm ISM} \ge 0.23]$, and let $y = 1$ if the \emph{Ph+Im} readout predicts the galaxy correctly, so that $p_1$ is the mean of $y$ within a cell. We fit 
  \begin{equation}
  P(y = 1 \mid a, b) = \sigma\!\left(\beta_0 + \beta_a a + \beta_b b + \beta_{ab}\, ab\right)
  \end{equation}
by unregularized maximum likelihood with the same Wald standard errors as above. If the two properties reflected a single mechanism, their effects would not combine independently, giving $\beta_{ab} \neq 0$. The interaction coefficient is instead consistent with zero, $\hat\beta_{ab} = +0.06 \pm 0.46$ ($95\%$ Wald CI; $p = 0.78$, two-sided Wald test), so each property raises the log-odds of a correct \emph{Ph+Im} prediction by an amount that does not depend on whether the other is high. The gains themselves are close to additive: relative to the (low, low) cell, high $m_{\rm fib,r}$ alone adds $2.5$ percentage points and high $\tau_{\rm ISM}$ alone adds $2.9$, a sum of $5.4$, against $5.6$ when both are high (Table~\ref{tab:2x2}).

\begin{table}[h]             
  \centering      
  \small
  \caption{Multimodal accuracy gain $\Delta = p_1 - p_2$ on the full held-out test set ($N = 13{,}551$), where $p_1 = P(\text{correct}\,|\,\emph{Ph+Im})$ and $p_2 = P(\text{correct}\,|\,\emph{Ph})$ are the fractions of galaxies in a cell predicted correctly by each readout, stratified by median splits of $m_{\rm fib,r}$ (median $20.23$) and $\tau_{\rm ISM}$ (median $0.23$). The $95\%$ CIs are computed within each cell as $\pm 1.96\sqrt{(p_1(1-p_1) + p_2(1-p_2))/N}$.}                   
  \label{tab:2x2}
  \begin{tabular}{lcc}   
  \toprule & low $\tau_{\rm ISM}$ & high $\tau_{\rm ISM}$ \\
  \midrule low $m_{\rm fib,r}$ & $+2.2\% \pm 0.7\%$ ($N{=}3{,}965$) & $+5.1\% \pm 1.1\%$ ($N{=}2{,}810$) \\
  high $m_{\rm fib,r}$ & $+4.7\% \pm 1.1\%$ ($N{=}2{,}810$) & $+7.8\% \pm 1.2\%$ ($N{=}3{,}966$) \\  \bottomrule                                   \end{tabular}
\end{table} 

\section{Layerwise probing and block-resolved causal localization}
\label{app:layerwise-probing}                              

This appendix supports Sec.~\ref{sec:which-block}. Appendix~\ref{app:layerwise-probes} gives the layerwise linear-probe protocol behind Fig.~\ref{fig:fig3} and the complete probe results (Fig.~\ref{fig:app-layerwise-full}). Appendix~\ref{app:block-causal} gives the block-resolved activation patching that confirms the block-1 localization by causal intervention.

\subsection{Layerwise linear probing}
\label{app:layerwise-probes}

\paragraph{Methodology.} For each input configuration (\emph{Ph} and \emph{Ph+Im}) and each of the 14 probed encoder stages, namely the token embedding (the encoder input, i.e.\ the output of \texttt{aion\_model.embed\_inputs}), the output of each of the 12 transformer blocks, and the output of the final encoder layer-norm, we run the frozen AION-1 encoder forward on both the $54{,}204$-galaxy training split, on which each probe is fit, and the $13{,}551$-galaxy held-out split, on which it is evaluated (Sec.~\ref{sec:setup}). Per-block and post-encoder-norm activations are captured by registering forward hooks on \texttt{aion\_model.encoder[i]} for $i = 0, \ldots, 11$ and on \texttt{aion\_model.encoder\_norm}; the token-embedding stage is captured directly as the output of \texttt{aion\_model.embed\_inputs}. At each stage, the per-token $768$-dimensional activations are mean-pooled along the token axis to a single $768$-dimensional vector per galaxy; at the token-embedding stage, where the sequence is padded to a fixed length, the encoder attention mask restricts the mean to the real tokens. For each (stage, property) pair we drop activation features whose training-split standard deviation is below $10^{-6}$ (near-constant features, which standardization cannot scale), standardize the remainder with \texttt{sklearn.preprocessing.StandardScaler} fit on the training split, and fit a ridge probe via \texttt{sklearn.linear\_model.RidgeCV} with a 13-point logarithmic $\alpha$ grid on $[10^{-1}, 10^{5}]$ and scikit-learn's closed-form efficient leave-one-out cross-validation (\texttt{cv=None}). Probes are restricted per target to galaxies on which the property is finite. Reported $R^2$ values are computed by \texttt{sklearn.metrics.r2\_score} on the held-out split. 

\begin{figure}[h]
      \centering
      \includegraphics[width=\textwidth]{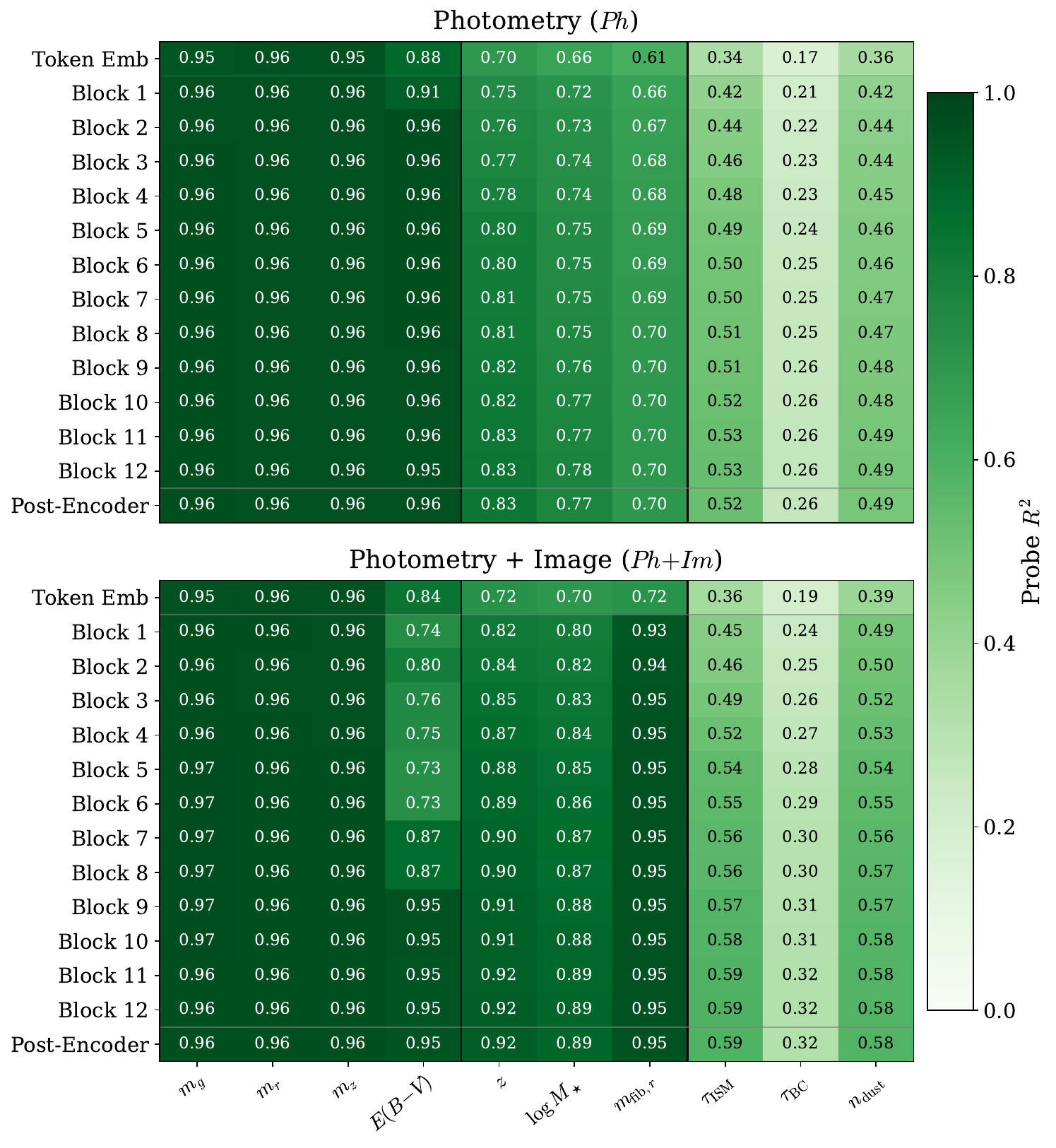}
      \caption{Layerwise linear-probe $R^2$ of the 10 galaxy properties of Fig.~\ref{fig:fig3} (columns) at all 14 encoder stages (rows), namely the token embedding (``Token Emb''), the outputs of each of the 12 transformer blocks, and the post-encoder representation. Columns are grouped (vertical lines) into three families: properties determined directly by the photometry tokens ($m_g$, $m_r$, $m_z$, $E(B{-}V)$); properties whose decodability the \emph{Ph+Im} configuration improves (redshift $z$, stellar mass $\log M_\star$, fiber $r$-magnitude $m_{\rm fib,r}$); and the weakly decodable dust-related parameters ($\tau_{\rm ISM}$, $\tau_{\rm BC}$, $n_{\rm dust}$). Top: \emph{Ph}; bottom: \emph{Ph+Im}.}

      \label{fig:app-layerwise-full}
  \end{figure}
  
\paragraph{Block structure and the attention/MLP split.} Each AION-1 encoder block is a standard pre-norm transformer block, $x \leftarrow x + \mathrm{Attn}(\mathrm{LN}_1(x))$ followed by $x \leftarrow x + \mathrm{MLP}(\mathrm{LN}_2(x))$, so the MLP sub-layer acts on the residual stream after the attention output has been added, and each block-$i$ stage above is the residual stream after both sub-layers. Probing the stream between the two sub-layers of block~1 as well places the block-1 redshift gain under \emph{Ph+Im} mainly in the attention sub-layer: $R^2(z)$ rises from $0.72$ at the token embedding to $0.79$ after block-1 attention and to $0.82$ after its MLP. 
  
\paragraph{Final-stage decodability.} The columns of Fig.~\ref{fig:app-layerwise-full} fall into three families, and the post-encoder (last) row summarizes how linearly decodable each is at the end of the encoder. First, the properties passed into the model directly as photometry tokens are recovered almost perfectly under both input configurations: the broadband magnitudes $m_g$, $m_r$, $m_z$ reach $R^2 = 0.96$, and $E(B{-}V)$ reaches $0.96$ (\emph{Ph}) and $0.95$ (\emph{Ph+Im}). Second, properties that are not photometry tokens but can be constrained by them, redshift, stellar mass, and $m_{\rm fib,r}$, are decodable from the post-encoder representation under \emph{Ph} ($R^2 = 0.83$, $0.77$, $0.70$, respectively) and improve under \emph{Ph+Im} ($0.92$, $0.89$, $0.95$, respectively); these are the properties whose decodability the image modality improves. Third, the dust parameters $\tau_{\rm ISM}$, $\tau_{\rm BC}$, and $n_{\rm dust}$ are neither photometry tokens nor well constrained by them and remain weakly decodable at the post-encoder in both configurations ($0.52$, $0.26$, $0.49$, respectively, under \emph{Ph}; $0.59$, $0.32$, $0.58$, respectively, under \emph{Ph+Im}). 

\paragraph{The token embedding is already informative.} Before any transformer block has run, the token embedding produced by AION-1's per-modality codecs and input embeddings is already an approximately linear encoding of the inputs, and several galaxy properties are linearly decodable from it. At the token-embedding stage (Fig.~\ref{fig:app-layerwise-full}, top rows), $m_g$, $m_r$, and $m_z$ are decodable at $R^2 \in [0.95, 0.96]$ under both input configurations: redshift at $R^2 = 0.70$ (\emph{Ph}) and $0.72$ (\emph{Ph+Im}), $\log M_\star$ at $0.66$ (\emph{Ph}) and $0.70$ (\emph{Ph+Im}), and $m_{\rm fib,r}$ at $0.61$ (\emph{Ph}) and $0.72$ (\emph{Ph+Im}).         

\paragraph{Gradual increase in decodability after block~1.} For properties whose decodability is not saturated at the token embedding, $R^2$ increases with depth after the first encoder block (Fig.~\ref{fig:app-layerwise-full}). Under \emph{Ph+Im}, $R^2(z)$ increases from $0.82$ at the output of block~1 to $0.92$ at the post-encoder, $R^2(\log M_\star)$ from $0.80$ to $0.89$, $R^2(\tau_{\rm ISM})$ from $0.45$ to $0.59$, $R^2(\tau_{\rm BC})$ from $0.24$ to $0.32$, and $R^2(n_{\rm dust})$ from $0.49$ to $0.58$, with no single block contributing more than $0.03$.

\subsection{Block-resolved activation patching}
\label{app:block-causal}

\paragraph{Clean and corrupted runs.} Block-resolved activation patching compares a clean forward pass against a corrupted one in which each test galaxy keeps its own photometry tokens but its image tokens are replaced with those of another test galaxy. The replacement is a one-position circular shift of the randomly ordered test set: each galaxy receives the image tokens of the preceding galaxy and the first galaxy receives those of the last, so no galaxy keeps its own image. The same assignment is used for both interventions below. Using another galaxy's image rather than zeros or a mean image preserves the marginal distribution of the image input \citep{wang2022ioi}, while the joint photometry--image distribution is not preserved. On the full test set, the two reference values are $R^2 = 0.931$ (clean) and $R^2 = -0.127$ (corrupted). On the most spatially extended quartile, defined throughout as the top quartile of $\Delta m_r \equiv m_{\rm fib,r} - m_r$ ($\Delta m_r \ge 1.890$, $N = 3{,}388$), they are $0.936$ and $-0.406$. Both corrupted values are far below the \emph{Ph} $R^2 = 0.869$. The readout does not revert to a photometry-only prediction when the image is replaced: its prediction depends on the substituted image.
  
\paragraph{Block contributions and the two interventions.} The clean and corrupted runs are the two reference runs for this test, which asks whether a single block's contribution is sufficient to restore the clean $R^2$ or necessary to maintain it. We construct hybrid forward passes that follow one reference run except at a single block, whose contribution is taken from the other run. Write block $\ell$'s contribution to the residual stream as $\Delta_\ell = \mathrm{out}_\ell - \mathrm{in}_\ell$, where $\mathrm{in}_\ell$ and $\mathrm{out}_\ell$ are the block's input and output. In \emph{denoising} (clean $\to$ corrupted patching, a test of sufficiency; \citealp{heimersheim2024patching}) we run the corrupted pass and overwrite block $\ell$'s output with $\mathrm{in}^{\rm corrupt}_\ell + \Delta^{\rm clean}_\ell$, injecting the block's clean contribution, recorded from the clean run, onto the corrupted stream. In \emph{noising} (corrupted $\to$ clean patching, a test of necessity) we run the clean pass and overwrite block $\ell$'s output with $\mathrm{in}^{\rm clean}_\ell + \Delta^{\rm corrupt}_\ell$, using the contribution recorded from the corrupted run. In both cases blocks after $\ell$ compute normally on the modified stream, and the trained \emph{Ph+Im} redshift readout is re-run unchanged. Each intervention is applied to one block at a time. We report the restored fraction $f_\ell = (R^2_{{\rm denoise},\ell} - R^2_{\rm corrupt}) / (R^2_{\rm clean} - R^2_{\rm corrupt})$ and the removed fraction $d_\ell = (R^2_{\rm clean} - R^2_{{\rm noise},\ell}) / (R^2_{\rm clean} - R^2_{\rm corrupt})$. Both fractions are normalized by the clean-minus-corrupted contrast $R^2_{\rm clean} - R^2_{\rm corrupt} = 1.06$, not by the multimodal advantage $R^2_{\rm Ph+Im} - R^2_{\rm Ph} = 0.061$; $f_\ell = 1$ means one block's clean contribution is sufficient to recover the full contrast and $d_\ell = 1$ that corrupting it alone is sufficient to remove it. Intervals are galaxy-paired percentile bootstraps as in Appendix~\ref{app:bootstrap}.

\begin{table}[h]
  \centering
  \caption{Per-block noising and denoising results on the held-out test set. $d_\ell$ is the fraction of $R^2_{\rm clean} - R^2_{\rm corrupt}$ removed by noising block $\ell$ alone; $f_\ell$ is the fraction restored by denoising block $\ell$ alone.}
  \label{tab:block-causal}
  \begin{tabular}{lcc}
  \toprule
  Block & removed $d_\ell$ & restored $f_\ell$ \\
  \midrule
  1  & $0.987\,[0.987, 0.988]$ & $0.9999\,[0.9998, 1.0000]$ \\
    2  & $0.036\,[0.033, 0.039]$ & $0.539\,[0.529, 0.549]$ \\ 
    3  & $0.015\,[0.014, 0.017]$ & $0.433\,[0.424, 0.442]$ \\
    4  & $0.010\,[0.009, 0.011]$ & $0.316\,[0.308, 0.325]$ \\
    5  & $0.012\,[0.011, 0.014]$ & $0.261\,[0.253, 0.270]$ \\
    6  & $0.007\,[0.006, 0.008]$ & $0.162\,[0.155, 0.170]$ \\
    7  & $0.008\,[0.007, 0.010]$ & $0.201\,[0.192, 0.209]$ \\
    8  & $0.015\,[0.013, 0.016]$ & $0.266\,[0.258, 0.274]$ \\
    9  & $0.024\,[0.022, 0.026]$ & $0.340\,[0.331, 0.349]$ \\
    10 & $0.015\,[0.013, 0.016]$ & $0.268\,[0.261, 0.277]$ \\
    11 & $0.013\,[0.011, 0.015]$ & $0.256\,[0.249, 0.263]$ \\
    12 & $0.023\,[0.021, 0.025]$ & $0.324\,[0.317, 0.331]$ \\
  \bottomrule
  \end{tabular}
\end{table}

\paragraph{Per-block results.} Table~\ref{tab:block-causal} reports both fractions for every block. Denoising and noising give the same result for block~1 and different results for later blocks. Block~1 is both sufficient ($f_1 = 0.9999\,[0.9998, 1.0000]$, and $0.9998\,[0.9997, 0.9999]$ on the most spatially extended quartile) and necessary ($d_1 = 0.987$), so the probe-based localization of Sec.~\ref{sec:which-block} is also supported by intervention. Noising a later block removes little of the contrast ($d_\ell \le 0.036$ for $\ell \ge 2$). Measured against the multimodal advantage, however, these decreases are not negligible: noising block~2 alone lowers $R^2$ from $0.931$ to $0.892$, a decrease equal to $62\%$ of the advantage, and noising any one of blocks $3$--$12$ gives a decrease equal to $12$--$42\%$ of it. Denoising a later block restores more of the contrast ($f_\ell = 0.16$--$0.54$ for $\ell \ge 2$) but leaves $R^2$ below the \emph{Ph} value of $0.869$ in every case, so no later block alone recovers any of the multimodal advantage. Later blocks therefore make non-zero contributions, but block~1 is the only block whose contribution is on its own both necessary and sufficient.

\section{Spatial extent experiments}
\label{app:extent}

This appendix contains the methods and full results behind Sec.~\ref{sec:spatial-extent}. Appendix~\ref{app:image-counterfactuals} defines the image counterfactuals. Appendix~\ref{app:head1-probes} reports the head-wise linear probes, zero-ablation, and attention and image-token analyses for the twelve block-1 heads. Appendix~\ref{app:head1-patching} describes the matched-redshift activation patching on Head~1, and Appendix~\ref{app:extent-erasure} the linear erasure of spatial extent from the residual stream.

\subsection{Image counterfactuals}
\label{app:image-counterfactuals}

\paragraph{Construction.} Each counterfactual replaces the image in each of the four bands ($g$, $r$, $i$, $z$) for every test galaxy and leaves the photometry tokens unchanged, so the input differs from the clean \emph{Ph+Im} input only in the image tokens. The \emph{compact template} is a two-dimensional Gaussian with $\sigma = 1.5$ pixels, approximating a point source, centered on the $160 \times 160$ frame at pixel coordinates $(79.5, 79.5)$. The \emph{extended blur} is the original band image convolved with an isotropic two-dimensional Gaussian kernel with $\sigma = 6$ pixels, with zero padding outside the frame.
  
\paragraph{Flux matching.} The image codec reads only the central $96 \times 96$ pixels of the frame. On the test set, a median $76\%$ of a cutout's $r$-band flux lies inside this crop (5th percentile $18\%$), so matching flux over the full frame would not match the flux the codec receives: a compact template matched over the full frame has a median $1.3$ times the original crop flux (95th percentile $5.4$ times). We therefore rescale each counterfactual, per band, so that its flux inside the crop equals that of the original image. Table~\ref{tab:image-counterfactuals} reports the per-quintile $R^2$ under both counterfactuals, alongside the clean \emph{Ph+Im} and \emph{Ph} values. If each counterfactual is instead rescaled so that its total flux over the full $160 \times 160$ frame equals that of the original image, $R^2$ differs from the values in Table~\ref{tab:image-counterfactuals} by at most $0.022$ in every $\Delta m_r$ quintile, and the trends across quintiles are unchanged.

\begin{table}[h]
  \centering
  \caption{Redshift $R^2$ on the held-out test set per $\Delta m_r$ quintile, for the \emph{Ph} readout and for the \emph{Ph+Im} readout with the original image (clean), the compact template, and the extended blur. $\Delta m_r \equiv m_{\rm fib,r} - m_r$ increases with the fraction of $r$-band light outside the $1.5''$ DESI fiber, so larger values correspond to more spatially extended galaxies. Intervals are $95\%$ paired bootstrap intervals (Appendix~\ref{app:bootstrap}). These are the values plotted in Fig.~\ref{fig:fig4}a.}
  \label{tab:image-counterfactuals}
  \resizebox{\textwidth}{!}{%
  \begin{tabular}{lcccc}
  \toprule
  $\Delta m_r$ quintile & \emph{Ph} & \emph{Ph+Im} clean & Compact template & Extended blur \\
  \midrule
  Q1 $[0.271, 1.076]$ & $0.815\,[0.788, 0.837]$ & $0.906\,[0.892, 0.918]$ & $0.671\,[0.645, 0.696]$ & $0.540\,[0.496, 0.579]$ \\
  Q2 $[1.076, 1.334]$ & $0.875\,[0.858, 0.889]$ & $0.922\,[0.908, 0.933]$ & $0.620\,[0.593, 0.646]$ & $0.738\,[0.709, 0.765]$ \\
  Q3 $[1.334, 1.614]$ & $0.880\,[0.858, 0.897]$ & $0.922\,[0.901, 0.938]$ & $0.561\,[0.531, 0.588]$ & $0.758\,[0.722, 0.788]$ \\
  Q4 $[1.614, 2.006]$ & $0.859\,[0.838, 0.876]$ & $0.923\,[0.908, 0.936]$ & $0.458\,[0.430, 0.482]$ & $0.788\,[0.763, 0.808]$ \\
  Q5 $[2.006, 6.614]$ & $0.835\,[0.816, 0.852]$ & $0.934\,[0.916, 0.948]$ & $0.309\,[0.272, 0.345]$ &
  $0.793\,[0.765, 0.818]$ \\
  \bottomrule
  \end{tabular}}
\end{table}

\subsection{Per-head analyses of block~1}
\label{app:head1-probes}

\paragraph{Linear probes on individual heads.} We fit a linear probe to the output of each attention head of block~1 separately. Heads are numbered $1$ to $12$, corresponding to head indices $0$ to $11$ of \texttt{aion\_model.encoder[0].attn}. The output of a head is $AV$, the product of its post-softmax attention weights $A$ and its value vectors $V$. The twelve head outputs are concatenated along the feature dimension, with Head~$k$ occupying features $64(k-1)+1$ through $64k$, and the output projection \texttt{aion\_model.encoder[0].attn.proj} maps this $768$-dimensional concatenation back to the residual stream. We use $AV$ directly, before concatenation and projection. For the \emph{Ph+Im} configuration, a head's output is a $585 \times 64$ matrix per galaxy ($9$ photometry and $576$ image tokens), which we mean-pool over tokens to a $64$-dimensional feature vector. Standardization on the training split, the ridge probe and its regularization grid, the restriction to galaxies with a finite target, and evaluation on the $13{,}551$-galaxy held-out split are as in Appendix~\ref{app:layerwise-probes}.

Column (A) of Table~\ref{tab:head-table} reports the probe $R^2$ of all twelve heads on eight targets: the three properties whose decodability the image modality substantially improves in Fig.~\ref{fig:fig3} ($z$, $m_{\rm fib,r}$, $\log M_\star$), the spatial extent proxy $\Delta m_r$, three standard photometric-redshift indicators (the colors $g{-}r$ and $r{-}z$ and the $r$-band surface brightness $\mu_r$), and the dust attenuation $\tau_{\rm ISM}$. The surface brightness $\mu_r$ is available only for a Galaxy Zoo-matched subset, so its probes are fit and evaluated on that subset ($4{,}679$ training and $1{,}183$ held-out galaxies). Intervals on individual probe $R^2$ values resample the held-out galaxies with the probe held fixed (Appendix~\ref{app:bootstrap}). Head~1 ranks first on seven of the eight targets, all except $\tau_{\rm ISM}$, which Head~3 leads. For each of these seven targets, Head~1's margin over the second-ranked head, with a $95\%$ paired bootstrap interval that resamples the same galaxies for both heads, ranges from $0.051\,[0.038, 0.064]$ on $\log M_\star$ to $0.251\,[0.236, 0.264]$ on $m_{\rm fib,r}$, and every interval excludes zero. The second-ranked head varies across these seven targets (Heads~3, 5, 7, and~9), so no other head is consistently second.

\begin{table}[h]
  \centering
  \newcommand{\rci}[3]{\begin{tabular}{@{}c@{}}#1\\[-2pt]{\scriptsize $[#2, #3]$}\end{tabular}}
  \caption{Linear probing and zero-ablation of the twelve block-1 heads with the \emph{Ph+Im} configuration. \textbf{(A)} Probe $R^2$ on each head's mean-pooled output for eight targets, with $95\%$ bootstrap intervals. \textbf{(B)} Change in the redshift $R^2$ of the \emph{Ph+Im} readout ($\times 10^{-3}$) when the head's output is zeroed, on the full held-out test set and in the most dust-attenuated $\tau_{\rm ISM}$ quintile, with $95\%$ paired bootstrap intervals. Bold marks the highest $R^2$ in each column of (A) and the largest decrease in each column of (B).}
  \label{tab:head-table}
  \textbf{(A) Probe $R^2$ on head output}\par\smallskip 
  \setlength{\tabcolsep}{3pt}
  \resizebox{\textwidth}{!}{%
  \begin{tabular}{l cccccccc}
  \toprule
  Block-1 head & $z$ & $m_{\rm fib,r}$ & $\Delta m_r$ & $\mu_r$ & $g{-}r$ & $r{-}z$ & $\log M_\star$ & $\tau_{\rm ISM}$ \\
  \midrule
  Head~1 & \rci{\textbf{0.697}}{0.687}{0.707} & \rci{\textbf{0.879}}{0.866}{0.892} & \rci{\textbf{0.767}}{0.750}{0.783} & \rci{\textbf{0.817}}{0.769}{0.854} & \rci{\textbf{0.929}}{0.924}{0.934} & \rci{\textbf{0.878}}{0.869}{0.886} & \rci{\textbf{0.646}}{0.633}{0.658} & \rci{0.193}{0.180}{0.206} \\
  Head~2 & \rci{0.475}{0.462}{0.488} & \rci{0.493}{0.474}{0.512} & \rci{0.602}{0.582}{0.620} & \rci{0.466}{0.408}{0.520} & \rci{0.583}{0.571}{0.595} & \rci{0.523}{0.510}{0.537} & \rci{0.420}{0.403}{0.437} & \rci{0.105}{0.094}{0.117} \\
  Head~3 & \rci{0.570}{0.558}{0.583} & \rci{0.629}{0.615}{0.642} & \rci{0.448}{0.428}{0.467} & \rci{0.457}{0.399}{0.508} & \rci{0.639}{0.628}{0.649} & \rci{0.673}{0.661}{0.684} & \rci{0.587}{0.572}{0.602} & \rci{\textbf{0.264}}{0.246}{0.280} \\
  Head~4 & \rci{0.514}{0.501}{0.526} & \rci{0.627}{0.611}{0.641} & \rci{0.548}{0.527}{0.568} & \rci{0.536}{0.485}{0.581} & \rci{0.538}{0.525}{0.550} & \rci{0.528}{0.514}{0.541} & \rci{0.470}{0.453}{0.487} & \rci{0.183}{0.168}{0.197} \\
  Head~5 & \rci{0.591}{0.580}{0.602} & \rci{0.546}{0.527}{0.565} & \rci{0.643}{0.623}{0.662} & \rci{0.509}{0.460}{0.553} & \rci{0.798}{0.791}{0.805} & \rci{0.715}{0.704}{0.726} & \rci{0.521}{0.506}{0.536} & \rci{0.115}{0.105}{0.125} \\
  Head~6 & \rci{0.566}{0.554}{0.578} & \rci{0.608}{0.591}{0.624} & \rci{0.508}{0.486}{0.528} & \rci{0.542}{0.493}{0.587} & \rci{0.690}{0.679}{0.700} & \rci{0.661}{0.648}{0.673} & \rci{0.561}{0.545}{0.577} & \rci{0.200}{0.185}{0.214} \\
  Head~7 & \rci{0.594}{0.581}{0.606} & \rci{0.602}{0.587}{0.616} & \rci{0.471}{0.450}{0.490} & \rci{0.454}{0.402}{0.502} & \rci{0.709}{0.699}{0.718} & \rci{0.704}{0.692}{0.715} & \rci{0.595}{0.579}{0.610} & \rci{0.242}{0.226}{0.259} \\
  Head~8 & \rci{0.570}{0.558}{0.582} & \rci{0.621}{0.607}{0.635} & \rci{0.499}{0.480}{0.519} & \rci{0.511}{0.455}{0.560} & \rci{0.654}{0.644}{0.663} & \rci{0.674}{0.663}{0.685} & \rci{0.586}{0.571}{0.601} & \rci{0.256}{0.240}{0.272} \\
  Head~9 & \rci{0.585}{0.573}{0.597} & \rci{0.611}{0.596}{0.625} & \rci{0.616}{0.595}{0.635} & \rci{0.577}{0.532}{0.618} & \rci{0.714}{0.703}{0.723} & \rci{0.661}{0.649}{0.673} & \rci{0.566}{0.550}{0.582} & \rci{0.175}{0.162}{0.188} \\
  Head~10 & \rci{0.541}{0.529}{0.554} & \rci{0.575}{0.559}{0.591} & \rci{0.585}{0.567}{0.602} & \rci{0.373}{0.296}{0.441} & \rci{0.589}{0.577}{0.600} & \rci{0.595}{0.580}{0.608} & \rci{0.513}{0.496}{0.529} & \rci{0.183}{0.168}{0.197} \\
  Head~11 & \rci{0.521}{0.508}{0.533} & \rci{0.599}{0.584}{0.613} & \rci{0.460}{0.440}{0.479} & \rci{0.462}{0.410}{0.510} & \rci{0.594}{0.582}{0.606} & \rci{0.599}{0.586}{0.612} & \rci{0.530}{0.513}{0.547} & \rci{0.190}{0.175}{0.203} \\
  Head~12 & \rci{0.538}{0.525}{0.551} & \rci{0.624}{0.610}{0.637} & \rci{0.482}{0.463}{0.500} & \rci{0.473}{0.418}{0.522} & \rci{0.573}{0.561}{0.585} & \rci{0.583}{0.569}{0.595} & \rci{0.523}{0.506}{0.539} & \rci{0.224}{0.208}{0.240} \\
  \bottomrule
  \end{tabular}}
  \par\bigskip
  \textbf{(B) Ablation $\Delta R^2$ ($\times 10^{-3}$), head zeroed}\par\smallskip
  \setlength{\tabcolsep}{6pt}
  \footnotesize
  \begin{tabular}{l cc}
  \toprule
  Block-1 head & Full test set & Most dust-attenuated $\tau_{\rm ISM}$ quintile \\
  \midrule
  Head~1 & $-0.47\,[-0.82, -0.18]$ & $-1.81\,[-3.22, -0.55]$ \\
  Head~2 & $-0.66\,[-1.29, -0.14]$ & $-1.37\,[-4.33, +1.11]$ \\
  Head~3 & $\mathbf{-52.39\,[-56.03, -49.34]}$ & $\mathbf{-108.44\,[-122.52, -94.96]}$ \\
  Head~4 & $-0.03\,[-0.07, +0.02]$ & $-0.16\,[-0.38, +0.01]$ \\
  Head~5 & $-0.02\,[-0.33, +0.27]$ & $+0.43\,[-0.72, +1.45]$ \\
  Head~6 & $-0.00\,[-0.02, +0.01]$ & $-0.03\,[-0.09, +0.02]$ \\
  Head~7 & $-0.89\,[-1.73, +0.01]$ & $+2.76\,[+0.03, +5.69]$ \\
  Head~8 & $-0.02\,[-0.08, +0.05]$ & $-0.38\,[-0.59, -0.18]$ \\
  Head~9 & $-2.73\,[-3.37, -2.15]$ & $-8.27\,[-10.79, -5.95]$ \\
  Head~10 & $-0.03\,[-0.05, -0.01]$ & $-0.08\,[-0.14, -0.01]$ \\
  Head~11 & $-2.81\,[-3.50, -2.12]$ & $-9.38\,[-12.18, -6.69]$ \\
  Head~12 & $-1.32\,[-1.81, -0.85]$ & $-3.19\,[-5.17, -1.29]$ \\
  \bottomrule
  \end{tabular}
\end{table}

\paragraph{Zero-ablation of individual heads.}  We zero-ablate each attention head of block~1 separately and measure the resulting change in the redshift $R^2$ of the \emph{Ph+Im} readout. To ablate Head~$k$, we set features $64(k-1)+1$ through $64k$ of the concatenated head output to zero at all $585$ token positions, immediately before the output projection \texttt{aion\_model.encoder[0].attn.proj} is applied. This removes the head's additive contribution to the residual stream and leaves the other eleven heads of block~1 and the trained readout unchanged.

Column (B) of Table~\ref{tab:head-table} reports the resulting change in redshift $R^2$, in units of $10^{-3}$, relative to the \emph{Ph+Im} readout, which reaches $R^2 = 0.931\,[0.924, 0.936]$ on the full held-out test set, with $95\%$ paired bootstrap intervals over $B = 1{,}000$ resamples (Appendix~\ref{app:bootstrap}). Ablating Head~3 decreases $R^2$ by $52.4 \times 10^{-3}$, every other head changes it by less than $3 \times 10^{-3}$, and Head~1 changes it by $-0.47\,[-0.82, -0.18] \times 10^{-3}$. Columns (A) and (B) rank the heads differently: Head~1 ranks first on seven of the eight probe targets but seventh of twelve by ablation, while Head~3 ranks first on $\tau_{\rm ISM}$, sixth on $z$, and last on $\Delta m_r$ among the probes, and first by ablation. The second ablation column, restricted to the most dust-attenuated $\tau_{\rm ISM}$ quintile, is analyzed in Appendix~\ref{app:head3-robustness}.
 
Because Head~1 is the head whose output best encodes $\Delta m_r$, we also evaluate its ablation within each $\Delta m_r$ quintile ($N \approx 2{,}710$ each), with $95\%$ paired bootstrap intervals over $B = 2{,}000$ resamples. From the least to the most spatially extended quintile, $\Delta R^2$ is $-0.60\,[-2.50, +0.74]$, $-0.01\,[-0.56, +0.54]$, $-0.37\,[-0.95, +0.26]$, $-0.40\,[-1.20, +0.31]$, and $-1.15\,[-1.88, -0.40]$, in units of $10^{-3}$. Every change is below $2 \times 10^{-3}$ in magnitude, and only the interval for the most spatially extended quintile excludes zero.

\paragraph{Attention map grouping and image-token PCA.} The attention map of Fig.~\ref{fig:fig4}b uses Head~1's attention weights $A$, and the principal-component analysis of Fig.~\ref{fig:fig4}c uses its output $AV$ (both defined above), each computed on a single \emph{Ph+Im} forward pass over the first $200$ galaxies of the held-out split. To assign each image token a brightness, we take the central $96 \times 96$ pixels that the image codec tokenizes (Appendix~\ref{app:redshift_pred}), average each $4 \times 4$-pixel block corresponding to one token, and average over the four bands. Tokens in the two outermost rows or columns of the $24 \times 24$ grid ($176$ tokens) are labeled \emph{border}; the remaining interior tokens are split within each galaxy at the $25$\textsuperscript{th} and $75$\textsuperscript{th} percentiles of that galaxy's interior-token brightness distribution into \emph{faint}, \emph{medium}, and \emph{bright}. With each of the $9$ photometry tokens as its own group, this gives $13$ groups. Attention is then averaged within each of the $13$ query groups and $13$ key groups, and across the $200$ galaxies. Because the groups contain different numbers of tokens, each cell reports the mean attention placed on a single key token of its column group rather than a sum, so rows do not sum to one. Summed over all key tokens of each modality, Head~1's image queries place a mean of $0.37\%$ of their attention on the nine photometry keys ($[10\textsuperscript{th}, 90\textsuperscript{th}] = [0.35\%, 0.42\%]$ across galaxies), and its photometry queries place $90.4\%$ ($[90.1\%, 90.8\%]$) on the $576$ image keys. In the photometry-query rows of Fig.~\ref{fig:fig4}b, each photometry key therefore receives more attention than any single image key, but the image keys together receive most of it.

For Fig.~\ref{fig:fig4}c, we run a separate PCA within each galaxy across the $576$ rows of Head~1's $(576, 64)$ image-token output, and correlate the per-token PC0 score with that token's Euclidean distance from the center of the $24 \times 24$ grid, taken as the continuous coordinate $(11.5, 11.5)$. Across the $200$ galaxies, the first component explains a mean of $84.1\%\,[83.6\%, 84.5\%]$ of the variance (median $84.5\%$, $[10\textsuperscript{th}, 90\textsuperscript{th}] = [80.4\%, 87.6\%]$), and $|\mathrm{corr}(\mathrm{PC0}, \mathrm{radius})|$ has mean $0.739\,[0.737, 0.741]$, median $0.741$, and a $[10\textsuperscript{th}, 90\textsuperscript{th}]$ range of $[0.719, 0.758]$.

\paragraph{Redundancy across heads.} The probes in Fig.~\ref{fig:fig4}d use the per-head mean-pooled outputs and the probe protocol above, with $\Delta m_r$ as the target, but concatenate the $64$-dimensional outputs of several heads before the probe is fit. A joint probe on all twelve heads reaches $R^2 = 0.891\,[0.878, 0.902]$. For each head, we also fit a leave-one-head-out probe on the concatenated outputs of the other eleven heads. Leaving out any head other than Head~1 reduces $R^2$ by at most $0.006$ relative to the joint probe. Leaving out Head~1 reduces it by $0.120$, to $0.771\,[0.756, 0.786]$, which is comparable to the $0.767\,[0.750, 0.783]$ of the probe on Head~1 alone. On their own, the twelve heads reach $\Delta m_r$ probe $R^2$ between $0.448$ and $0.767$ (Table~\ref{tab:head-table}).

\subsection{Matched-redshift activation patching on Head 1}
\label{app:head1-patching}

Zero-ablation replaces Head~1's output with values it never takes on real inputs. Activation patching instead replaces it with Head~1's output on another galaxy, so the activations stay within the range the readout was trained on. If the readout used the spatial extent information in Head~1's output, then giving a high-$\Delta m_r$ galaxy the Head~1 output of a low-$\Delta m_r$ galaxy at the same redshift should move its predicted redshift toward the low-$\Delta m_r$ galaxy's prediction.
  
\paragraph{Pair construction.} We take the lowest ($\Delta m_r \le 1.076$) and highest ($\Delta m_r \ge 2.006$) $\Delta m_r$ quintiles of the held-out test set, $2{,}711$ galaxies each, referred to as low-$\Delta m_r$ and high-$\Delta m_r$. We write a pair as recipient$\leftarrow$donor: in the high$\leftarrow$low direction, each high-$\Delta m_r$ galaxy is a recipient that receives Head~1's output from a low-$\Delta m_r$ donor, and the low$\leftarrow$high direction reverses the roles. Within each direction, pairs are formed greedily over a random permutation of the recipients. For each recipient in turn, we step outward from its position in the list of donors sorted by $z_{\rm spec}$, one position at a time on both sides, skipping donors already used, and stop at the first step at which an unused donor within $|\Delta z_{\rm spec}| \le 0.005$ has been found. Each donor is used at most once. Because donors are not evenly spaced in redshift, the chosen donor is the closest unused donor for $73\%$ of pairs; otherwise, it is at most $0.002$ farther in $z_{\rm spec}$ than the closest unused donor (median $7 \times 10^{-5}$). The two directions are matched independently, giving $N = 1{,}583$ high$\leftarrow$low and $N = 1{,}584$ low$\leftarrow$high pairs, with mean $|\Delta z_{\rm spec}|$ of $3.7 \times 10^{-4}$ and $5.6 \times 10^{-4}$. The remaining $42\%$ of recipients in each direction are unmatched; apart from $37$ high$\leftarrow$low recipients with no donor within tolerance, every unmatched recipient had donors within tolerance that had already been paired with earlier recipients.
  
\paragraph{Patching procedure.} Patching intervenes at the same point as the zero-ablation of Appendix~\ref{app:head1-probes}, features $1$ through $64$ of the concatenated head output before the output projection, but writes the donor's values instead of zeros. For each pair, we run the frozen \emph{Ph+Im} encoder on the donor and record Head~1's output at all $585$ token positions. We then run it on the recipient with Head~1's output replaced by the donor's at every position, leaving the other eleven heads unchanged, and apply the trained \emph{Ph+Im} readout. The photometry positions are patched as well as the image positions because Head~1's photometry queries place $90.4\%$ of their attention on the image keys (Appendix~\ref{app:head1-probes}), so its output at those positions also carries image information. As a check, patching each of the first $320$ held-out galaxies with its own Head~1 output reproduces its unpatched prediction exactly.

\paragraph{Patching regression.} For each pair, the patched shift $\Delta z_{\rm pred} = z_{\rm pred}^{\rm patched} - z_{\rm pred}^{\rm clean}$ is the change in the recipient's predicted redshift, and the prediction gap $\Delta z_{\rm gap} = z_{\rm pred,\,donor}^{\rm clean} - z_{\rm pred,\,recipient}^{\rm clean}$ is the difference between the donor's and the recipient's unpatched predictions. Within each direction, we regress $\Delta z_{\rm pred}$ on $\Delta z_{\rm gap}$ by ordinary least squares. A slope of $1$ means that patching Head~1 alone moves the recipient's prediction to the donor's, and a slope of $0$ means that the patched prediction does not move with the gap. Intervals are $95\%$ percentile bootstrap intervals that resample the matched pairs of each direction with replacement, keeping each pair's $\Delta z_{\rm pred}$ and $\Delta z_{\rm gap}$ together, with the pairing held fixed (Appendix~\ref{app:bootstrap}).
  
\paragraph{Result.} The high$\leftarrow$low direction gives a slope of $0.008\,[0.004, 0.011]$ ($N = 1{,}583$) and the low$\leftarrow$high direction $0.003\,[-0.001, 0.007]$ ($N = 1{,}584$). The mean patched shift is $+0.0009\,[0.0008, 0.0010]$ and $-0.0004\,[-0.0005, -0.0003]$, against root-mean-square prediction gaps of $0.035$ and $0.036$.
  
\subsection{Erasing the linear spatial extent direction from the residual stream}
\label{app:extent-erasure}

Zero-ablating a head tests whether spatial extent is carried by that component; erasing a direction tests whether it is carried by a linear subspace of the residual stream. We use LEACE (least-squares concept erasure; \citealp{belrose2023leace}). For a scalar target, the LEACE eraser is the rank-1 affine map 
  \[
    r(x) = x - \frac{\sigma\,\sigma^\top \Sigma^{-1} (x - \mu)}{\sigma^\top \Sigma^{-1} \sigma},
  \]
where $\mu$ and $\Sigma$ are the mean and covariance of the activations and $\sigma$ is their cross-covariance with the target. The erased activations $r(x)$ have zero cross-covariance with the target, which implies that no linear predictor of the target from $r(x)$ attains a lower mean squared error than a constant prediction. Among affine maps with this property, LEACE minimizes the expected squared Euclidean distance $\mathbb{E}\,\lVert r(x) - x \rVert^2$. Erasing $\Delta m_r$ therefore removes it from every linear decoder, not only from the probe of Appendix~\ref{app:head1-probes}. We regularize $\Sigma$ by adding $10^{-3}$ times its mean eigenvalue to the diagonal; the zero-cross-covariance property holds for any such regularization.
  
\paragraph{Fitting and applying the eraser.} We estimate $\mu$, $\Sigma$, and $\sigma$ on the mean-pooled block-1 output activations (Block~1 post-MLP in Table~\ref{tab:extent-erasure-sweep}) of the training split under the unmodified forward pass, with $\Delta m_r$ as the target. During the forward pass, $r$ is applied to the block-1 output at every token position, and the remaining blocks run unchanged. Because $r$ is affine, mean-pooling the erased token activations gives the same vector as erasing the mean-pooled activation (maximum absolute deviation $3 \times 10^{-6}$), so the guarantee above applies to the mean-pooled representation on which the probes are fit; it does not constrain the individual token activations.
  
\paragraph{Verification and control.} On the training split, the largest absolute entry of the cross-covariance between the erased activations and $\Delta m_r$ is $2 \times 10^{-15}$. On the held-out split, a $\Delta m_r$ probe refit on the erased block-1 output reaches $R^2 = -2.5 \times 10^{-5}\,[-4.7 \times 10^{-4}, -1.5 \times 10^{-7}]$, against $0.903\,[0.889, 0.914]$ before erasure. As a control, we fit the same eraser to a single random permutation of the $\Delta m_r$ labels, which removes a direction of the same rank that is not determined by $\Delta m_r$; the probe then reaches $0.902\,[0.888, 0.912]$, so the loss of decodability is specific to the $\Delta m_r$ direction. With either eraser in place, we probe every subsequent stage of the encoder with the protocol of Appendix~\ref{app:layerwise-probes} (Table~\ref{tab:extent-erasure-sweep}). Intervals resample the held-out galaxies with each probe held fixed, and differences between conditions use paired resamples (Appendix~\ref{app:bootstrap}).

\paragraph{$\Delta m_r$ is decodable again downstream.} Immediately after block~2's attention sub-layer, the $\Delta m_r$ probe reaches $R^2 = 0.909\,[0.895, 0.920]$ on the erased stream against $0.912\,[0.898, 0.923]$ on the clean stream (paired decrease $0.003\,[0.001, 0.005]$), and at the post-encoder stage $0.913\,[0.899, 0.924]$ against $0.936\,[0.922, 0.946]$. Within one attention sub-layer of the intervention, $\Delta m_r$ is again nearly as linearly decodable as without erasure, so the rank-1 projection does not remove the information from which a linear $\Delta m_r$ representation can be formed. These measurements concern decodability only; they do not establish whether the encoder recomputes the erased representation or forms a different one.
  
\paragraph{Redshift decodability recovers only partly.} A redshift probe on the post-encoder representation reaches $R^2 = 0.865\,[0.857, 0.871]$ with the eraser in place, against $0.918\,[0.911, 0.924]$ on the clean stream and $0.908\,[0.901, 0.914]$ with the control eraser. The decrease of $0.054\,[0.050, 0.058]$ under erasure is $0.043\,[0.040, 0.047]$ larger than the control's $0.011\,[0.009, 0.012]$. Redshift decodability therefore recovers only partly after the erasure: the decrease is $0.105\,[0.099, 0.111]$ at block~1's output and $0.054\,[0.050, 0.058]$ at the post-encoder stage.

\begin{table}[!htbp]
  \centering
  \caption{Per-stage effect of the block-1 spatial extent eraser: held-out probe $R^2$ for $\Delta m_r$ and for redshift at every encoder stage under the clean forward pass, with the rank-1 LEACE eraser applied at block~1's output, and with the shuffled-label control eraser. Stages are those of Appendix~\ref{app:layerwise-probes}; ``post-attn'' is the residual stream between a block's attention and MLP sub-layers. $^{\ast}$Site of the intervention; the first two rows precede it and are identical across conditions. Each entry gives the probe $R^2$ with its $95\%$ bootstrap interval below (probe held fixed; Appendix~\ref{app:bootstrap}). $^{\dagger}$Rounds to zero at three decimals: $-2.5 \times 10^{-5}\,[-4.7 \times 10^{-4}, -1.5 \times 10^{-7}]$.}
  \label{tab:extent-erasure-sweep}
  \newcommand{\rci}[3]{\begin{tabular}{@{}c@{}}#1\\[-2pt]{\scriptsize $[#2, #3]$}\end{tabular}}
  \small
  \setlength{\tabcolsep}{3pt}
  \begin{tabular}{l ccc ccc}
  \toprule
  & \multicolumn{3}{c}{$\Delta m_r$ probe $R^2$} & \multicolumn{3}{c}{redshift probe $R^2$} \\
  \cmidrule(lr){2-4} \cmidrule(lr){5-7}
  Stage & clean & erased & shuffled & clean & erased & shuffled \\
  \midrule
Token Emb & \rci{0.577}{0.558}{0.594} & \rci{0.577}{0.558}{0.594} & \rci{0.577}{0.558}{0.594} & \rci{0.722}{0.712}{0.731} & \rci{0.722}{0.712}{0.731} & \rci{0.722}{0.712}{0.731} \\
Block 1 post-attn & \rci{0.892}{0.879}{0.903} & \rci{0.892}{0.879}{0.903} & \rci{0.892}{0.879}{0.903} & \rci{0.795}{0.786}{0.803} & \rci{0.795}{0.786}{0.803} & \rci{0.795}{0.786}{0.803} \\
Block 1 post-MLP$^{\ast}$ & \rci{0.903}{0.889}{0.914} & \rci{$0.000^{\dagger}$}{-0.000}{-0.000} & \rci{0.902}{0.888}{0.912} & \rci{0.820}{0.812}{0.828} & \rci{0.715}{0.704}{0.724} & \rci{0.819}{0.811}{0.827} \\
Block 2 post-attn & \rci{0.912}{0.898}{0.923} & \rci{0.909}{0.895}{0.920} & \rci{0.912}{0.898}{0.922} & \rci{0.823}{0.815}{0.831} & \rci{0.817}{0.809}{0.825} & \rci{0.823}{0.815}{0.831} \\
Block 2 post-MLP & \rci{0.915}{0.902}{0.926} & \rci{0.910}{0.897}{0.921} & \rci{0.914}{0.900}{0.924} & \rci{0.836}{0.828}{0.843} & \rci{0.817}{0.808}{0.824} & \rci{0.830}{0.822}{0.837} \\
Block 3 post-attn & \rci{0.917}{0.903}{0.928} & \rci{0.911}{0.897}{0.921} & \rci{0.916}{0.902}{0.927} & \rci{0.840}{0.832}{0.848} & \rci{0.821}{0.813}{0.829} & \rci{0.837}{0.829}{0.844} \\
Block 3 post-MLP & \rci{0.922}{0.908}{0.932} & \rci{0.912}{0.898}{0.923} & \rci{0.920}{0.906}{0.930} & \rci{0.850}{0.842}{0.857} & \rci{0.827}{0.819}{0.834} & \rci{0.844}{0.836}{0.851} \\
Block 4 post-attn & \rci{0.925}{0.912}{0.935} & \rci{0.914}{0.901}{0.925} & \rci{0.922}{0.909}{0.932} & \rci{0.860}{0.853}{0.868} & \rci{0.833}{0.825}{0.840} & \rci{0.851}{0.844}{0.858} \\
Block 4 post-MLP & \rci{0.928}{0.915}{0.938} & \rci{0.916}{0.903}{0.926} & \rci{0.927}{0.915}{0.937} & \rci{0.866}{0.858}{0.873} & \rci{0.837}{0.830}{0.845} & \rci{0.857}{0.850}{0.865} \\
Block 5 post-attn & \rci{0.930}{0.917}{0.940} & \rci{0.917}{0.903}{0.927} & \rci{0.929}{0.916}{0.938} & \rci{0.880}{0.873}{0.886} & \rci{0.846}{0.838}{0.853} & \rci{0.873}{0.866}{0.880} \\
Block 5 post-MLP & \rci{0.932}{0.919}{0.942} & \rci{0.919}{0.905}{0.929} & \rci{0.930}{0.917}{0.940} & \rci{0.883}{0.876}{0.890} & \rci{0.849}{0.841}{0.856} & \rci{0.877}{0.869}{0.883} \\
Block 6 post-attn & \rci{0.932}{0.919}{0.943} & \rci{0.918}{0.904}{0.928} & \rci{0.930}{0.917}{0.940} & \rci{0.886}{0.879}{0.893} & \rci{0.848}{0.841}{0.855} & \rci{0.880}{0.873}{0.887} \\
Block 6 post-MLP & \rci{0.934}{0.921}{0.944} & \rci{0.918}{0.904}{0.928} & \rci{0.931}{0.918}{0.942} & \rci{0.890}{0.883}{0.897} & \rci{0.852}{0.844}{0.859} & \rci{0.884}{0.877}{0.890} \\
Block 7 post-attn & \rci{0.935}{0.921}{0.944} & \rci{0.916}{0.903}{0.927} & \rci{0.931}{0.917}{0.941} & \rci{0.891}{0.884}{0.898} & \rci{0.851}{0.844}{0.858} & \rci{0.885}{0.877}{0.891} \\
Block 7 post-MLP & \rci{0.936}{0.923}{0.945} & \rci{0.916}{0.902}{0.927} & \rci{0.932}{0.919}{0.942} & \rci{0.895}{0.888}{0.901} & \rci{0.853}{0.846}{0.860} & \rci{0.889}{0.882}{0.895} \\
Block 8 post-attn & \rci{0.936}{0.922}{0.945} & \rci{0.916}{0.902}{0.927} & \rci{0.932}{0.918}{0.942} & \rci{0.896}{0.889}{0.902} & \rci{0.853}{0.846}{0.860} & \rci{0.890}{0.883}{0.896} \\
Block 8 post-MLP & \rci{0.936}{0.923}{0.946} & \rci{0.916}{0.901}{0.926} & \rci{0.932}{0.918}{0.943} & \rci{0.901}{0.894}{0.907} & \rci{0.857}{0.850}{0.864} & \rci{0.894}{0.887}{0.900} \\
Block 9 post-attn & \rci{0.936}{0.923}{0.946} & \rci{0.916}{0.901}{0.927} & \rci{0.932}{0.918}{0.943} & \rci{0.902}{0.895}{0.908} & \rci{0.857}{0.850}{0.864} & \rci{0.894}{0.887}{0.901} \\
Block 9 post-MLP & \rci{0.936}{0.923}{0.946} & \rci{0.915}{0.901}{0.926} & \rci{0.933}{0.919}{0.943} & \rci{0.907}{0.900}{0.913} & \rci{0.858}{0.851}{0.865} & \rci{0.898}{0.891}{0.904} \\
Block 10 post-attn & \rci{0.936}{0.922}{0.946} & \rci{0.914}{0.900}{0.925} & \rci{0.933}{0.919}{0.943} & \rci{0.909}{0.902}{0.915} & \rci{0.859}{0.852}{0.866} & \rci{0.900}{0.893}{0.906} \\
Block 10 post-MLP & \rci{0.936}{0.923}{0.946} & \rci{0.915}{0.900}{0.926} & \rci{0.933}{0.919}{0.943} & \rci{0.912}{0.905}{0.918} & \rci{0.860}{0.853}{0.867} & \rci{0.903}{0.896}{0.909} \\
Block 11 post-attn & \rci{0.936}{0.922}{0.947} & \rci{0.915}{0.900}{0.926} & \rci{0.933}{0.919}{0.943} & \rci{0.913}{0.906}{0.919} & \rci{0.860}{0.853}{0.868} & \rci{0.904}{0.897}{0.910} \\
Block 11 post-MLP & \rci{0.936}{0.923}{0.946} & \rci{0.914}{0.899}{0.925} & \rci{0.933}{0.919}{0.943} & \rci{0.915}{0.908}{0.922} & \rci{0.862}{0.855}{0.869} & \rci{0.906}{0.900}{0.913} \\
Block 12 post-attn & \rci{0.936}{0.923}{0.946} & \rci{0.913}{0.899}{0.924} & \rci{0.933}{0.919}{0.943} & \rci{0.916}{0.909}{0.922} & \rci{0.862}{0.855}{0.869} & \rci{0.907}{0.900}{0.913} \\
Block 12 post-MLP & \rci{0.936}{0.922}{0.946} & \rci{0.914}{0.900}{0.925} & \rci{0.932}{0.918}{0.943} & \rci{0.918}{0.912}{0.924} & \rci{0.865}{0.857}{0.872} & \rci{0.908}{0.901}{0.914} \\
Post-Encoder & \rci{0.936}{0.922}{0.946} & \rci{0.913}{0.899}{0.924} & \rci{0.932}{0.918}{0.943} & \rci{0.918}{0.911}{0.924} & \rci{0.865}{0.857}{0.871} & \rci{0.908}{0.901}{0.914} \\
  \bottomrule
  \end{tabular}
\end{table}

\section{Dust attenuation experiments}
\label{app:head3}

This appendix contains the methods and full results behind Sec.~\ref{sec:dust-attenuation}. Appendix~\ref{app:head3-robustness} reports the ablation and sufficiency results for Head~3 and the robustness checks on the head ranking, and Appendix~\ref{app:slope-bias} decomposes the prediction error into attenuation bias and scatter. Appendices~\ref{app:head3-probes}--\ref{app:head3-cross-modal} evaluate four families of candidate operations for Head~3, and Appendix~\ref{app:head3-concat} tests whether Head~3's output closes the redshift probe gap. Appendix~\ref{app:head3-circuit} covers the model-wide head search and the downstream circuit analysis.

\subsection{Ablation, sufficiency, and robustness of the head ranking}
\label{app:head3-robustness}

\paragraph{Per-head zero-ablation.} Column (B) of Table~\ref{tab:head-table} reports the effect of zero-ablating each block-1 head (protocol in Appendix~\ref{app:head1-probes}) as $\Delta R^2$, the redshift $R^2$ with the head ablated minus that of the \emph{Ph+Im} readout. On the full held-out test set, zero-ablating Head~3 gives $\Delta R^2 = -0.052\,[-0.056, -0.049]$, against $-0.003\,[-0.004, -0.002]$ for the next-largest effect (Head~11). On the most dust-attenuated $\tau_{\rm ISM}$ quintile ($\tau_{\rm ISM} \geq 0.563$), where the \emph{Ph+Im} readout reaches $R^2 = 0.863\,[0.838, 0.886]$, zero-ablating Head~3 gives $\Delta R^2 = -0.108\,[-0.122, -0.095]$, more than eleven times the next-largest effect ($-0.009\,[-0.012, -0.007]$, Head~11).

\paragraph{Per-quintile results.} Table~\ref{tab:h3-quintiles} reports the redshift $R^2$ of the \emph{Ph} and \emph{Ph+Im} readouts in each $\tau_{\rm ISM}$ quintile, together with that of the \emph{Ph+Im} readout when Head~3 is zero-ablated (\emph{Head~3 zeroed}) and when the other eleven block-1 heads are mean-ablated so that Head~3 is the only active head in block~1 (\emph{Head~3 only}). The $\Delta R^2$ column, \emph{Head~3 zeroed} minus \emph{Ph+Im}, has $95\%$ paired bootstrap intervals over $10{,}000$ resamples of the galaxies within each quintile, and the other columns over $2{,}000$; all five $\Delta R^2$ intervals exclude zero, and the interval for the most dust-attenuated quintile overlaps no other.

\begin{table}[h]
  \centering
  \caption{Redshift $R^2$ per $\tau_{\rm ISM}$ quintile on the held-out test set for the
    \emph{Ph} and \emph{Ph+Im} readouts and for the \emph{Ph+Im} readout with the other eleven
    block-1 heads mean-ablated (\emph{Head~3 only}) and with Head~3 zero-ablated
    (\emph{Head~3 zeroed}), with $95\%$ paired bootstrap intervals. $\Delta R^2$ is
    \emph{Head~3 zeroed} minus \emph{Ph+Im}. These are the values plotted in
    Fig.~\ref{fig:fig5}a.}
  \label{tab:h3-quintiles}
  \resizebox{\textwidth}{!}{%
  \begin{tabular}{lccccr}
  \toprule
  $\tau_{\rm ISM}$ quintile & \emph{Ph} & \emph{Ph+Im} & Head~3 only & Head~3 zeroed & $\Delta R^2$ \\
  \midrule
  Q1 $[0.000, 0.088]$ & $0.910\,[0.894, 0.922]$ & $0.952\,[0.943, 0.960]$ & $0.947\,[0.937, 0.955]$ & $0.907\,[0.896, 0.917]$ & $-0.045\,[-0.052, -0.039]$ \\
  Q2 $[0.088, 0.167]$ & $0.914\,[0.901, 0.926]$ & $0.954\,[0.942, 0.964]$ & $0.951\,[0.939, 0.961]$ & $0.930\,[0.918, 0.939]$ & $-0.024\,[-0.028, -0.020]$ \\
  Q3 $[0.167, 0.314]$ & $0.874\,[0.855, 0.889]$ & $0.938\,[0.928, 0.947]$ & $0.936\,[0.925, 0.944]$ & $0.894\,[0.881, 0.904]$ & $-0.045\,[-0.052, -0.038]$ \\
  Q4 $[0.314, 0.563]$ & $0.806\,[0.779, 0.828]$ & $0.913\,[0.892, 0.928]$ & $0.912\,[0.889, 0.927]$ & $0.850\,[0.831, 0.865]$ & $-0.063\,[-0.072, -0.055]$ \\
  Q5 $[0.563, 3.000]$ & $0.774\,[0.751, 0.798]$ & $0.863\,[0.838, 0.886]$ & $0.863\,[0.839, 0.885]$ & $0.754\,[0.729, 0.779]$ & $-0.108\,[-0.122, -0.095]$ \\
  \bottomrule
  \end{tabular}}
\end{table}

\paragraph{Mean ablation and the sufficiency test.} Zero-ablation tests whether Head~3 is necessary for the multimodal advantage; the sufficiency test asks the converse, whether Head~3 alone among the block-1 heads is enough to reproduce the \emph{Ph+Im} readout's performance. We remove the other eleven heads and leave the rest of the encoder, including block~1's MLP, unchanged, so this is a test of sufficiency among the block-1 heads only. Zero-ablating eleven heads at once would move block~1's output far from the distribution that the later blocks receive, so we use mean ablation instead: each head's output is replaced by its per-position mean over a random subsample of $5{,}000$ training-split galaxies (seed $123$), a fixed vector that does not depend on the input galaxy. Leaving only Head~3 active and mean-ablating the other eleven gives $R^2 = 0.928\,[0.922, 0.934]$, against $0.931\,[0.924, 0.936]$ for the \emph{Ph+Im} readout and $0.869\,[0.861, 0.877]$ for \emph{Ph}. The paired gap to the \emph{Ph+Im} readout is $\Delta R^2 = -0.0025\,[-0.0035, -0.0015]$, so Head~3 alone recovers $96\%\,[94\%, 98\%]$ of the multimodal advantage of $0.061\,[0.055, 0.067]$. Per quintile, Head~3 alone recovers $87\%\,[81\%, 92\%]$, $94\%\,[88\%, 98\%]$, $96\%\,[92\%, 99\%]$, $99\%\,[96\%, 101\%]$, and $100\%\,[96\%, 104\%]$ of the multimodal advantage from Q1 to Q5. The fraction is largest in the two most dust-attenuated quintiles, where the multimodal advantage is largest, and there its interval includes $100\%$.

\paragraph{Robustness.} Three checks support reading Head~3's rank as a property of the model rather than of our particular intervention or data split.

First, Head~3's effect remains significant after correcting for the number of heads and evaluation sets tested. Ranking twelve heads on six evaluation sets (the full test set and the five $\tau_{\rm ISM}$ quintiles) amounts to $72$ tests of whether an ablation changes $R^2$, and with that many tests, some small effects appear significant by chance. We therefore test each $\Delta R^2$ against zero and correct for the number of tests. Each test's two-sided $p$-value is estimated from $10{,}000$ paired bootstrap resamples as twice the fraction of resamples in which $\Delta R^2$ has the opposite sign to the observed value, with a minimum of $10^{-4}$, the smallest value $10{,}000$ resamples can resolve. We then apply the Holm correction, which keeps the probability of any false positive among the $72$ tests at $5\%$. No resample gives Head~3 a $\Delta R^2 \ge 0$ in any evaluation set, so its $p$-value is $10^{-4}$ in all six, or $0.007$ after correction, and it is the only head that remains significant in all six. On the full test set, Heads~9, 10, 11, and 12 also remain significant, but the largest of their effects (Head~11, $\Delta R^2 = -0.003\,[-0.004, -0.002]$) is far smaller than Head~3's $-0.052\,[-0.056, -0.049]$.

Second, Head~3's rank does not depend on the readout's early-stopping criterion or data split. The readouts used throughout the paper are early-stopped on the test set; Appendix~\ref{app:redshift_pred} describes two retrained pairs of readouts that are instead early-stopped on a validation set held out from the training split, one on the original partition and one on a second partition. Their \emph{Ph+Im} readouts reach $R^2 = 0.930\,[0.924, 0.936]$ and $0.927\,[0.920, 0.934]$. Repeating the ablation of all twelve block-1 heads on each ranks Head~3 first, with $\Delta R^2 = -0.050\,[-0.053, -0.046]$ and $-0.053\,[-0.057, -0.050]$, while ablating Head~1 gives $-0.0008\,[-0.0012, -0.0005]$ and $-0.0006\,[-0.0009, -0.0002]$. Head~3's first-place rank and Head~1's near-zero effect therefore do not depend on the early-stopping criterion, the particular readout weights, or the particular test set.

Third, Head~3's effect does not depend on the choice of ablation. Replacing Head~3's output with its per-position mean rather than zero gives $R^2 = 0.832\,[0.825, 0.839]$, a change of $\Delta R^2 = -0.099\,[-0.103, -0.094]$, larger in magnitude than the $-0.052\,[-0.056, -0.049]$ that zeroing produces. The effect is therefore not an artifact of the zero reference lying off the distribution of the head's outputs; zeroing gives the smaller of the two estimates of Head~3's contribution.

\subsection{Decomposition of the prediction error into attenuation bias and scatter}
\label{app:slope-bias}

\paragraph{Decomposition.} For redshift predictions $z_{\rm pred}$ evaluated against spectroscopic redshifts $z_{\rm spec}$ on a set of galaxies, we fit $z_{\rm pred} = \alpha + \beta z_{\rm spec}$ by ordinary least squares. Because the residual of this fit is uncorrelated with $z_{\rm spec}$, the variance of the prediction error decomposes as
\begin{equation}
\mathrm{Var}(z_{\rm pred} - z_{\rm spec}) = (1-\beta)^2\,\mathrm{Var}(z_{\rm spec}) + \sigma^2_{\rm scatter},
\label{eq:slopebias}
\end{equation}
where $\sigma^2_{\rm scatter}$ is the variance of the fit residual. The first term is the attenuation-bias variance: a slope $\beta < 1$ compresses the predictions toward their mean, and the term vanishes only when $\beta = 1$. Here, attenuation is the statistical term for a slope shrunk toward zero, distinct from dust attenuation. The second term is the scatter about the fitted line. Intervals in this subsection use $1{,}000$ bootstrap resamples; for a difference between two readouts, both are evaluated on the same resampled galaxies (Appendix~\ref{app:bootstrap}).

\paragraph{Slope and bias.} All results in this subsection are for the most dust-attenuated $\tau_{\rm ISM}$ quintile ($N = 2{,}711$, $\tau_{\rm ISM} \geq 0.563$). Adding imaging reduces the attenuation bias: the slope rises from $\beta = 0.806\,[0.786, 0.826]$ for the \emph{Ph} readout to $0.867\,[0.851, 0.883]$ for the \emph{Ph+Im} readout, and the attenuation-bias variance falls to $0.47\,[0.38, 0.57]$ times its \emph{Ph} value. Zero-ablating Head~3 eliminates this reduction and raises the attenuation bias above its \emph{Ph} level: the slope falls to $\beta = 0.710\,[0.692, 0.727]$, below the \emph{Ph} slope, and the attenuation-bias variance rises to $2.2\,[1.9, 2.7]$ times its \emph{Ph} value, so high-redshift galaxies are underpredicted and low-redshift galaxies overpredicted more strongly than with photometry alone. The scatter, by contrast, remains slightly below that of the \emph{Ph} readout after ablation: $\sigma_{\rm scatter} = 0.039\,[0.037, 0.041]$ against $0.042\,[0.040, 0.044]$ (paired change $-0.003\,[-0.005, -0.001]$).
  
\subsection{Probes for expected physical parameters}
\label{app:head3-probes}

\paragraph{Hypothesis and motivation.} Zero-ablating Head~3 removes most of the multimodal advantage, and the effect is largest in the most dust-attenuated (highest $\tau_{\rm ISM}$) quintile (Appendix~\ref{app:head3-robustness}). The most direct explanation is that Head~3 encodes the physical quantities that make dust-attenuated galaxies hard to predict from photometry alone: the dust attenuation itself, or a dust-insensitive estimate of intrinsic luminosity. The second possibility is suggested by Head~3's attention to the WISE $W1$ and $W2$ tokens (Fig.~\ref{fig:fig5}b), since these infrared bands are standard stellar-mass indicators that are only weakly affected by dust. If either holds, the quantity should be more decodable from Head~3's output than from the other block-1 heads, and removing it from Head~3's output should remove Head~3's causal effect. We test these explanations with four analyses. All linear probes use the per-head protocol of Appendix~\ref{app:head1-probes}. 
  
\paragraph{Probes for the quantities in Head~3's ablation result.} We first probe each block-1 head for three quantities tied directly to Head~3's ablation result: the redshift $z$, which the readout predicts; $\tau_{\rm ISM}$, the property along which the ablation effect grows; and the multimodal correction $z_{\rm pred}^{\rm Ph+Im} - z_{\rm pred}^{\rm Ph}$, the per-galaxy difference between the two readouts' predictions (probes fit on training-split predictions and evaluated on the held-out split). For comparison, we also probe $\Delta m_r$, the spatial extent proxy. Head~3 does not stand out on redshift or on the correction, and leads on $\tau_{\rm ISM}$ only weakly (Table~\ref{tab:head-table}). Its redshift probe reaches $R^2 = 0.570\,[0.558, 0.583]$, in the middle of the twelve heads, and its $\Delta m_r$ probe ranks last. It ranks first on $\tau_{\rm ISM}$, but at $R^2 = 0.264\,[0.246, 0.280]$, against $0.589\,[0.568, 0.608]$ for a linear probe on the post-encoder representation. Every head decodes the multimodal correction only weakly, with $R^2$ from $0.044\,[0.036, 0.052]$ to $0.109\,[0.097, 0.121]$; Head~1 is highest and Head~3 reaches $0.062\,[0.052, 0.072]$. Zeroing Head~3 removes most of the multimodal advantage, yet the per-galaxy correction is no more decodable from Head~3's output than from that of Head~1, whose ablation has no measurable effect.

\paragraph{Probes for other candidate quantities.} We next probe each head's output mean-pooled over the $576$ image tokens only, rather than over all $585$ tokens as in Appendix~\ref{app:head1-probes}, for two further sets of candidates.  Surface brightness is an image-derived quantity used in photometric-redshift estimation, so we probe $r$-band surface brightness $\mu_r$ on a subset matched to Galaxy Zoo DECaLS ($N = 1{,}183$ test galaxies) and the inner-to-outer surface-brightness gradient. Head~3's attention to $W1$ and $W2$ suggests a luminosity or stellar-mass estimate based on these bands, so we probe the PROVABGS stellar mass $\log M_\star$, apparent $W1$ magnitude, and $W1{-}r$ color. Head~3 ranks tenth of twelve on $\mu_r$ ($R^2 = 0.456\,[0.399, 0.508]$, against $0.817\,[0.768, 0.854]$ for the best head) and tenth on the gradient ($0.168\,[0.155, 0.180]$, against $0.244\,[0.231, 0.257]$). On the full test set, it ranks third on $\log M_\star$ ($0.587\,[0.571, 0.601]$) and sixth on both $W1$ magnitude ($0.397\,[0.360, 0.438]$) and $W1{-}r$ ($0.283\,[0.228, 0.333]$). Within the most dust-attenuated quintile, the intervals are too wide to rank heads: Head~3 reaches $0.016\,[-0.068, 0.094]$, $0.323\,[0.217, 0.444]$, and $0.127\,[-0.186, 0.283]$ on the three targets, respectively, each interval overlapping that of the top head, and no head exceeds $R^2 = 0.04$ on stellar mass. Head~3 is not distinguished on any of these targets.
  
\paragraph{Nonlinear probes.} A linear probe can miss information that is encoded nonlinearly, so we also probe Head~3's image-token mean-pooled output for $\tau_{\rm ISM}$ with an MLP: one hidden layer of $256$ ReLU units, trained with Adam (learning rate $10^{-3}$, $L_2$ penalty $10^{-4}$, early stopping on a $10\%$ validation split with patience $10$, at most $300$ iterations, seed $42$), with a control trained on permuted targets. The MLP recovers more than a linear probe on the same features,  $R^2 = 0.397\,[0.379, 0.414]$ against $0.264\,[0.247, 0.279]$, and the control gives $-0.009\,[-0.012, -0.006]$, but it remains well below the $0.589$ of a linear probe on the post-encoder representation. The weak linear result is therefore not explained by a nonlinear encoding that this MLP can recover. Intervals use $B = 1{,}000$ resamples.

\paragraph{Erasing dust information from Head~3's output.} Weak decodability does not rule out that the little dust information in Head~3's output is what the readout uses. We therefore remove it with rank-1 LEACE erasure (Appendix~\ref{app:extent-erasure}), which removes the linearly decodable information about a target while changing the output as little as possible, and re-run the \emph{Ph+Im} readout. Unlike the spatial extent eraser of Appendix~\ref{app:extent-erasure}, this eraser is fit on Head~3's per-token outputs for $4{,}000$ training galaxies, because an eraser fit on the mean-pooled output distorts the individual token outputs enough that even its shuffled-label control degrades the prediction more than zero-ablating the whole head. After erasure, a probe on the mean-pooled output recovers $\tau_{\rm ISM}$ at only $R^2 = 0.014\,[0.010, 0.017]$, against $0.264\,[0.246, 0.280]$ before; a small residual remains because the eraser acts on each token separately while the probe reads their average. In the most dust-attenuated quintile, erasure lowers the readout's $R^2$ by no more than a control eraser fit to shuffled $\tau_{\rm ISM}$ labels (difference $-0.002\,[-0.007, +0.003]$). Erasing the multimodal correction $z_{\rm pred}^{\rm Ph+Im} - z_{\rm pred}^{\rm Ph}$ instead lowers it by $0.019\,[0.012, 0.026]$ more than the control, so erasure does change the prediction when the erased information is used. The dust information that probes find in Head~3's output is therefore not what the readout relies on. Intervals use $B = 1{,}000$ resamples, except the post-erasure probe $R^2$, which uses $B = 2{,}000$.

\subsection{Dereddening and $4000\,$\AA\ break tracking}
\label{app:head3-photometry-attn}

\paragraph{Hypothesis and motivation.} In Head~3's \emph{Ph+Im} attention (Fig.~\ref{fig:fig5}b), every photometry query attends most to the $g$ flux token and the $E(B{-}V)$ token. Two standard photometric-redshift operations would produce this pattern. The first is dereddening: $E(B{-}V)$ is the foreground reddening from Milky Way dust along the line of sight, and correcting the observed fluxes for it is a routine preprocessing step, so a head that combines the $g$ flux with $E(B{-}V)$ could be computing a dereddened flux. This correction concerns Galactic dust rather than the dust within the galaxy that $\tau_{\rm ISM}$ measures, so this hypothesis is suggested by the attention pattern only. The second is tracking the $4000\,$\AA\ break, the strongest redshift-dependent feature in the optical spectra of these galaxies: as redshift increases, the break moves through the $g$ band toward the $r$ band, so a head that tracks it should change its attention specifically between the $g$ and $r$ flux tokens. The break enters $r$ only near $z \approx 0.37$, beyond the redshifts of most galaxies in the sample (quintile medians $0.08$--$0.33$), so we test for the gradual shift between $g$ and $r$ as the break moves through $g$, not for a discontinuity at the transition.
  
\paragraph{Dereddening.} A dereddening head's output at the $g$ flux position should vary with $E(B{-}V)$ and should track the dereddened $g$ flux more closely than the observed one. For each of the nine photometry positions, we compute the principal components of Head~3's output across a random subset of $200$ held-out galaxies (those of Fig.~\ref{fig:fig5}b, for which Head~3's per-token outputs and attention maps were stored) and correlate each galaxy's score on the first principal component with $E(B{-}V)$ and with the observed and dereddened $g$ flux, the latter computed with the coefficients of \citet{schlafly2011measuring}. We find that neither prediction holds. At the $g$ flux position, the first principal component is uncorrelated with $E(B{-}V)$ (Pearson correlation $-0.03\,[-0.17, +0.10]$). It correlates equally with the observed and dereddened $g$ flux (correlations $0.369\,[0.301, 0.447]$ and $0.364\,[0.297, 0.447]$), as expected for any output because $E(B{-}V)$ is small for these galaxies (median $0.029$), so the two fluxes are nearly identical (correlation $0.999$ between them). At the $E(B{-}V)$ position, the first principal component correlates only weakly with $E(B{-}V)$ (correlation $0.28\,[0.16, 0.39]$). Correlation intervals resample the $200$ galaxies with the principal components held fixed.

\paragraph{$4000\,$\AA\ break tracking.} A break-tracking head should shift its attention between the $g$ and $r$ flux tokens as redshift increases, more than between other bands. For each of the $200$ galaxies, we compute the attention that the photometry and image queries place on each photometry token; for the image queries we average over the $576$ image tokens and express each photometry token's weight as a share of their total attention to the photometry tokens. We average these quantities within each redshift quintile and fit a line to the averages against each quintile's median redshift (intervals over $1{,}000$ galaxy resamples). The attention of every photometry token to the $g$ flux token increases with redshift, opposite to the prediction (slope $+0.21\,[+0.13, +0.27]$ per unit redshift for the $g$ flux token's attention to itself). The image tokens' attention is redistributed among the photometry tokens as redshift increases, but not from $g$ to $r$: the share of their photometry attention on the $g$ flux token decreases (slope $-0.11\,[-0.13, -0.08]$ per unit redshift, from $0.097$ in the lowest redshift quintile to $0.067$ in the highest), the share on the $r$ flux token shows no trend ($-0.04\,[-0.07, +0.00]$), and the shares on the $i$ and $W1$ tokens rise ($+0.10\,[+0.03, +0.16]$ and $+0.10\,[+0.04, +0.16]$). 
  
Because the attention results are not specific to the break, we also test whether Head~3 encodes break-related information. In broadband photometry, the break appears as a color with two components: its wavelength, which shifts with redshift and determines how $g{-}r$ varies with redshift, and its amplitude, which determines $g{-}r$ at fixed redshift. The first is equivalent to redshift itself, which Head~3 encodes no better than the other block-1 heads (Appendix~\ref{app:head3-probes}). For the second, we probe each head on the full test set for the residual $g{-}r$ color after subtracting a linear fit of $g{-}r$ on redshift (fit on the training split). This residual, how much redder or bluer a galaxy is than the average galaxy at its redshift, approximates the break amplitude but also reflects dust reddening. Head~3 ranks seventh of twelve ($R^2 = 0.435\,[0.421, 0.449]$, against $0.616\,[0.603, 0.629]$ for Head~1). Head~3 therefore encodes neither the redshifted wavelength of the break nor its amplitude more than the other block-1 heads do.
  
\subsection{Image-token spatial encoding}
\label{app:head3-image}
\paragraph{Hypothesis and motivation.}  Part of the improvement that images give photometric-redshift estimates has been attributed to information beyond integrated photometry: morphological parameters such as galaxy size, shape, and surface brightness~\citep{morphoz2020}, and the redshift information in individual pixels, which deep networks combine more effectively than integrated photometry does, reducing attenuation bias~\citep{moran2026regions}. Dust attenuation varies across a galaxy, so a head that encodes resolved color or brightness structure could correct for it. If Head~3 did, four predictions follow: its image-token outputs should vary across a galaxy with the local color or brightness, its ablation should matter most for galaxies whose color varies most across the image, its effect should be concentrated at the image tokens that cover the galaxy's light, and any dust information in its output should be concentrated at particular token positions.
  
\paragraph{Within-galaxy correlation with color and brightness maps.} To compare each image token's output with the color and brightness at that token's position, we compute these maps at the resolution of the token grid, as in Appendix~\ref{app:head1-probes}: the image codec encodes the central $96 \times 96$ pixels of each cutout as a $24 \times 24$ grid of image tokens, so each token corresponds to a $4 \times 4$-pixel block, and we average each map over that block. For each of the $200$ held-out galaxies of Fig.~\ref{fig:fig5}b, we compute the principal components of Head~3's outputs across the galaxy's $576$ image tokens and correlate each token's score on the first principal component with five maps of the same galaxy: brightness (the mean over the four bands), $g{-}r$ color, $g{-}z$ color, the local $g{-}z$ gradient (at each token, the largest absolute difference in $g{-}z$ from its four adjacent tokens, which is large where the color changes sharply, as at a dust lane), and distance from the image center. The mean absolute correlation over galaxies is at most $0.13$: $0.120\,[0.115, 0.124]$ for brightness, $0.126\,[0.117, 0.137]$ for $g{-}r$, $0.128\,[0.118, 0.138]$ for $g{-}z$, $0.087\,[0.078, 0.097]$ for the gradient, and $0.077\,[0.069, 0.084]$ for distance from the center. For comparison, the same analysis applied to Head~1 gives a mean absolute correlation of $0.739\,[0.737, 0.741]$ between the first-principal-component score and distance from the image center (Sec.~\ref{sec:extent-head1}), so the analysis does detect a dependence of a head's output on token position when one is present.

\paragraph{Ablation effect versus color variation across the galaxy.} For every galaxy in the test set, we compute $\sigma_{\rm resid}$, the standard deviation over the $576$ image tokens of the $g{-}z$ map after subtracting its galaxy mean, divide the test set into quintiles of $\sigma_{\rm resid}$, and compute Head~3's zero-ablation $\Delta R^2$ in each. The effect shrinks as color variation grows, from $\Delta R^2 = -0.084\,[-0.095, -0.074]$ in the lowest quintile ($\sigma_{\rm resid} \in [0.03, 0.38]$) to $-0.047\,[-0.056, -0.038]$ in the highest ($[0.55, 2.49]$; difference $+0.038\,[+0.024, +0.051]$), the opposite of the prediction. The lowest and highest quintiles have comparable mean $\tau_{\rm ISM}$, so the decline is not a consequence of the dependence of Head~3's effect on $\tau_{\rm ISM}$.

\paragraph{Token-subset ablation.} If Head~3's effect came from resolved structure in the galaxy, zero-ablating Head~3 only at the image tokens that cover the galaxy's light should reproduce most of its effect. We approximate these tokens in two ways, as the brightest quarter of each galaxy's image tokens and as the $112$ image tokens within $6$ tokens of the image center, and compare them with the complementary subsets. Each $\Delta R^2$ in the most dust-attenuated quintile is expressed as a fraction of the $-0.106\,[-0.119, -0.093]$ obtained by ablating all $576$ image tokens. The tokens covering the galaxy carry almost none of the effect: the brightest quarter accounts for $2.0\%\,[0.9\%, 3.2\%]$ and the central tokens for $3.0\%\,[1.9\%, 4.1\%]$. The remaining $464$ outer tokens account for $41.9\%\,[35.9\%, 47.2\%]$, the faintest quarter for $-0.8\%\,[-2.4\%, 0.6\%]$, and ablating Head~3 at the $9$ photometry tokens instead for $1.5\%\,[0.5\%, 2.8\%]$. The central and outer subsets together cover all image tokens, yet their separate ablations sum to $-0.048\,[-0.057, -0.038]$, less than half of the joint $-0.106$, so Head~3's effect depends on many image tokens jointly.
  
\paragraph{Per-token probes.}  Appendix~\ref{app:head3-probes} uses probes fit on Head~3's output mean-pooled over the image tokens, and mean pooling could dilute dust information present at only a few positions. We therefore fit a separate linear probe at each of the $576$ image-token positions; if such information existed, the best single-token probe should exceed the mean-pooled probe. It does not: for $\tau_{\rm ISM}$, the best token reaches $R^2 = 0.237\,[0.220, 0.253]$, below the $0.264\,[0.246, 0.280]$ of the mean-pooled probe, and for redshift, the best token reaches $0.557\,[0.546, 0.568]$ against $0.570\,[0.558, 0.583]$. The per-token probes do detect position-specific information when it is present: for $m_{\rm fib,r}$, which is measured in an aperture at the galaxy center, the best token, one of the four at the center of the token grid, reaches $0.701\,[0.687, 0.713]$, above the $0.629\,[0.615, 0.642]$ of the mean-pooled probe.
  
\subsection{Correction of photometry--image discrepancies}
\label{app:head3-cross-modal}
  
\paragraph{Hypothesis and motivation.} Head~3's image queries place a mean of $24.9\%\,[24.2\%, 25.5\%]$ of their attention on the nine photometry tokens, summed over key tokens as in Appendix~\ref{app:head1-probes} ($19.5\%$--$30.5\%$ between the $10$th and $90$th percentiles over the $200$ galaxies of Fig.~\ref{fig:fig5}b), against $0.37\%\,[0.37\%, 0.38\%]$ for Head~1. Through this attention, each of Head~3's image-token outputs includes a weighted sum of the photometry tokens' value vectors, so Head~3's output could correct the image-token representation of a galaxy where it is inconsistent with the photometry-token representation, or the reverse; here, inconsistent means that the two representations encode different values of the same property, such as the $g$-band magnitude. If Head~3's effect came from such a correction, its ablation should matter most for the galaxies whose photometry-token and image-token representations are most inconsistent.
  
\paragraph{Ablation effect versus photometry--image discrepancy.} We measure the inconsistency through the $g$-band magnitude $m_g$. We use a magnitude because both modalities measure it directly, as the $g$ flux token and as the $g$-band pixels, so the two representations should encode the same value. Two linear probes, fit on the training split, predict $m_g$ from the token embeddings that enter block~1, mean-pooled separately over the $9$ photometry tokens and the $576$ image tokens (held-out $R^2 = 0.974\,[0.973, 0.975]$ and $0.639\,[0.620, 0.657]$); because these embeddings precede block~1, the discrepancy is unaffected by ablating any block-1 head. After centering each probe's predictions on their test-set mean, the discrepancy is $D = \hat m_g^{\rm Ph} - \hat m_g^{\rm Im}$. We divide the test set into quintiles of $|D|$ and compute Head~3's zero-ablation $\Delta R^2$ in each. Instead of growing with $|D|$, Head~3's effect shrinks, from $\Delta R^2 = -0.066\,[-0.075, -0.056]$ in the lowest $|D|$ quintile to $-0.031\,[-0.038, -0.025]$ in the highest.  $|D|$ is also uncorrelated with $\tau_{\rm ISM}$ (Spearman $\rho = -0.009\,[-0.026, +0.007]$), so the decline is not a consequence of the dependence of Head~3's effect on $\tau_{\rm ISM}$. Because the image-token probe is less accurate than the photometry-token probe, $|D|$ largely reflects the error of the image-token probe and is therefore an approximate measure of the inconsistency between the two representations.

\subsection{Closing the redshift probe gap with Head~3's output}
\label{app:head3-concat}

\paragraph{Hypothesis and motivation.} The tests of Appendices~\ref{app:head3-probes}--\ref{app:head3-cross-modal} each assume a specific operation. This test assumes none: if Head~3's output contains the information that distinguishes the \emph{Ph+Im} representation from the \emph{Ph} representation, in a form that a probe can extract, then adding Head~3's output to the \emph{Ph} representation should close a substantial fraction of the redshift probe gap. We quantify this as
  \begin{equation}
  f = \frac{R^2_{\rm concat} - R^2_{\rm Ph}}{R^2_{\rm Ph+Im} - R^2_{\rm Ph}},
  \end{equation}
the fraction of the probe gap closed, where $R^2_{\rm Ph}$ and $R^2_{\rm Ph+Im}$ are redshift probes on the mean-pooled post-encoder representations of the two inputs, and $R^2_{\rm concat}$ is a probe of the same type on the \emph{Ph} representation concatenated with Head~3's output. The probes are fit on frozen features, and the \emph{Ph+Im} readout is unchanged.
  
\paragraph{Mean-pooled output.} We first concatenate the $768$-dimensional \emph{Ph} representation with Head~3's $64$-dimensional output mean-pooled over the image tokens. With linear probes (protocol of Appendix~\ref{app:head1-probes}), $R^2_{\rm Ph} = 0.828\,[0.819, 0.836]$, $R^2_{\rm Ph+Im} = 0.918\,[0.911, 0.924]$, and $R^2_{\rm concat} = 0.832\,[0.824, 0.840]$, so $f = 0.050\,[0.034, 0.066]$. In the most dust-attenuated quintile, where Head~3 alone recovers the full multimodal advantage (Sec.~\ref{sec:head3-necessary}), $f = 0.013\,[-0.006, 0.032]$. With the MLP of Appendix~\ref{app:head3-probes} in place of linear probes, $f = 0.056\,[0.011, 0.096]$.
  
\paragraph{Token-resolved output.} Mean pooling discards which token each output comes from, so we repeat the linear test with the $576$ image-token outputs kept separate, each projected onto its leading $8$ or $16$ principal directions, fit on the image-token outputs of $3{,}000$ training galaxies ($96\%$ and $99.6\%$ of per-token variance). The token-resolved output alone gives redshift $R^2 = 0.762\,[0.752, 0.771]$ with $8$ directions, against $0.570\,[0.558, 0.583]$ for the mean-pooled output, so mean pooling does discard redshift information. Concatenated with the \emph{Ph} representation, the token-resolved output closes $f = 0.099\,[0.060, 0.134]$ of the gap with $8$ directions and $0.102\,[0.056, 0.146]$ with $16$, a rise over the mean-pooled output of $0.049\,[0.014, 0.083]$ with $8$ directions. In every case, Head~3's output closes at most about a tenth of the probe gap, although Head~3 is necessary for most of the multimodal advantage. These probes test only whether the information can be extracted from Head~3's output directly; they do not exclude a contribution of Head~3's output through computation in later blocks.

\subsection{Model-wide search and downstream circuit}
\label{app:head3-circuit}
\paragraph{Hypothesis and motivation.} We zero-ablated only the twelve heads of block~1, which leaves two alternatives open: heads in later blocks could have effects comparable to Head~3's, or Head~3's effect on the prediction could be mediated mainly by one or a few later heads whose input includes Head~3's output. We test the first with an attribution screen over all $144$ heads followed by zero-ablation of the heads it flags, and the second with two mediation tests on the later heads that are the strongest candidate mediators: single-edge path patching and ablation with frozen downstream heads. We label head $h$ of block $b$ as B$b$H$h$, with both indices starting at $1$.
  
\paragraph{Attribution screen over all 144 heads.} Zero-ablating each of the $144$ heads separately would require $144$ forward passes over the test set, so we first rank the heads with gradient$\times$activation attribution~\citep{syed2024attribution}, a first-order approximation to the change in the prediction under zero-ablation. For each head, we multiply its $64$-dimensional output, taken before the attention output projection, elementwise by the gradient of the predicted redshift with respect to that output, and sum over dimensions and over all $585$ token positions; one backward pass per galaxy gives the scores of all $144$ heads. The negative of a head's score approximates the change in each galaxy's predicted redshift when the head is zero-ablated; adding it to the clean predictions and recomputing $R^2$ gives the head's predicted $\Delta R^2$. On block~1, where every head has also been zero-ablated, the predicted and measured $\Delta R^2$ rank the twelve heads similarly (Spearman $\rho = 0.958\,[0.790, 0.993]$), and Head~3's predicted $\Delta R^2 = -0.056\,[-0.060, -0.053]$ is close to its measured $-0.052\,[-0.056, -0.049]$. We flag every head outside block~1 whose predicted $|\Delta R^2|$ in the most dust-attenuated $\tau_{\rm ISM}$ quintile exceeds $0.0083$, the $|\Delta R^2|$ measured by zero-ablating Head~9 on the same galaxies. Among the block-1 heads other than Head~3, this is the second-largest effect in that quintile, after Head~11's $0.0094$ (Table~\ref{tab:head-table}). Five heads are flagged: B5H4, B2H3, B5H10, B2H12, and B8H7.
  
\paragraph{Zero-ablation of the flagged heads.} We zero-ablate each flagged head with the protocol used for the block-1 heads. In the most dust-attenuated quintile, the measured effects follow the order predicted by the screen, and four of the five exceed the $0.0083$ threshold; we refer to these four as the confirmed heads. The exception is B8H7, at $\Delta R^2 = -0.0075\,[-0.0089, -0.0062]$. Over the full test set, every effect is negative with an interval excluding zero, but all are small, ranging from $-0.0046\,[-0.0053, -0.0040]$ (B5H4) to $-0.0022\,[-0.0026, -0.0019]$ (B8H7), an order of magnitude below Head~3's $-0.052\,[-0.056, -0.049]$. The screen therefore finds no head outside block~1 with an effect comparable to Head~3's.
  
\paragraph{Single-edge path patching.} A later head with a small effect of its own could still mediate a large part of Head~3's effect, because its input includes Head~3's output. Path patching~\citep{wang2022ioi} tests this one head at a time. For a later head X, we run the model on the clean input but compute X's queries, keys, and values from a residual stream from which Head~3's output has been removed; every other head receives the clean residual stream, so only the edge from Head~3 to X is cut. AION-1 computes the queries, keys, and values of all heads in a block with one linear projection, so we compute this projection on both residual streams and replace X's slice of the clean result with its slice of the ablated result; with no ablation, this procedure reproduces the clean predictions to within $1.3 \times 10^{-7}$. The edge fraction $E$ is the $\Delta R^2$ over the full test set from cutting the edge, divided by the $\Delta R^2$ from zero-ablating Head~3, so $E = 1$ would mean that all of Head~3's effect is mediated by X.
  
We cut the edges from Head~3 to the four confirmed heads (B2H3, B2H12, B5H4, B5H10); to B9H1 and B10H10, which the screen did not flag in the most dust-attenuated quintile but which rank among the five heads with the largest predicted $|\Delta R^2|$ over the full test set; and, as controls, to the two heads in each of blocks $2$, $5$, $9$, and $10$ with the smallest predicted $|\Delta R^2|$ in the most dust-attenuated quintile. Every edge fraction is small: the largest are $E = 0.071\,[0.059, 0.083]$ for B5H4, $0.047\,[0.034, 0.061]$ for B2H12, and $0.040\,[0.030, 0.051]$ for B2H3, against at most $0.0016\,[0.0004, 0.0029]$ for a control edge. Cutting an edge nonetheless changes the output of the receiving head substantially. Measured as the Frobenius norm of the change divided by that of the clean output, the change is $0.79$--$1.05$ for the four confirmed heads and $0.49$--$0.79$ for the eight control heads (each interval narrower than $\pm 0.01$). The later heads' outputs therefore depend strongly on Head~3's output, but cutting any single edge reproduces at most $7\%$ of Head~3's effect on the prediction.
  
\paragraph{Ablation with frozen downstream heads.} Path patching cuts one edge at a time while every other head still receives Head~3's output, so it does not measure what the four confirmed heads mediate jointly, including through each other. To measure this, we zero-ablate Head~3 while fixing the four confirmed heads to their clean-run outputs, so that the ablation cannot propagate through them. The mediated fraction $M = 1 - \Delta R^2(\text{ablated, frozen}) / \Delta R^2(\text{ablated})$ is the fraction of the ablation effect that freezing removes: $M = 1$ if the frozen heads mediate all of Head~3's effect and $M = 0$ if they mediate none of it. Freezing heads changes the computation even when they do not mediate Head~3's effect, so as a reference, we also freeze the four path-patching control heads in blocks~2 and~5, the blocks containing the confirmed heads. In the most dust-attenuated quintile, freezing the four confirmed heads gives $M = 0.030\,[-0.006, 0.064]$ and freezing the four control heads gives $M = -0.041\,[-0.051, -0.032]$; the mediated fraction of the confirmed heads is thus small and of similar magnitude to that of heads not selected as candidate mediators. Neither test finds that a single later head, or the four confirmed heads jointly, mediates most of Head~3's effect. Both tests cover attention heads only; mediation by many heads with small individual contributions or by the MLP layers, and a direct effect of Head~3's output through the residual stream on the readout, are not excluded.



\end{document}